\documentclass[12pt]{iopart}

\usepackage{iopams}  
\usepackage{url}            
\usepackage{booktabs}       
\usepackage{nicefrac}       
\usepackage{graphicx}
\usepackage{pdflscape}
\usepackage{xfrac}

\begin{document}

\title[Loss shaping enhances Eventprop learning]{Loss shaping enhances exact gradient learning with Eventprop in Spiking Neural Networks}

\author{Thomas Nowotny$^{*}$}
\address{School of Engineering and Informatics, University of Sussex, Brighton, BN1 9QJ, UK}
\ead{t.nowotny@sussex.ac.uk}
\author{James P. Turner}
\address{Information \& Communication Technologies, Imperial College London, London, SW7 2AZ, UK}
\ead{james.turner@imperial.ac.uk}
\author{James C. Knight}
\address{School of Engineering and Informatics, University of Sussex, Brighton, BN1 9QJ, UK}
\ead{j.c.knight@sussex.ac.uk}

\vspace{10pt}
\begin{indented}
\item[]$^*$Corresponding author
\end{indented}

\vspace{10pt}
\begin{indented}
\item[]December 2024 
\end{indented}

\begin{abstract}
Event-based machine learning promises more energy-efficient AI on future neuromorphic hardware. 
Here, we investigate how the recently discovered Eventprop algorithm for gradient descent on exact gradients in spiking neural networks can be scaled up to challenging keyword recognition benchmarks. 
We implemented Eventprop in the GPU-enhanced Neural Networks framework and used it for training recurrent spiking neural networks on the Spiking Heidelberg Digits and Spiking Speech Commands datasets. 
We found that learning depended strongly on the loss function and extended Eventprop to a wider class of loss functions to enable effective training. 
We then tested a large number of data augmentations and regularisations as well as exploring different network structures; and heterogeneous and trainable timescales. We found that when combined with two specific augmentations, the right regularisation and a delay line input, Eventprop networks with one recurrent layer achieved state-of-the-art performance on Spiking Heidelberg Digits and good accuracy on Spiking Speech Commands. 
In comparison to a leading surrogate-gradient-based SNN training method, our GeNN Eventprop implementation is 3$\times$ faster and uses 4$\times$ less memory.
This work is a significant step towards a low-power neuromorphic alternative to current machine learning paradigms.
\end{abstract}

%
%
%
%
%

\newcommand{\taumi}{\tau_{\text{mem},i}}
\newcommand{\tausi}{\tau_{\text{syn},i}}

\section{Introduction}
Modern deep neural networks need kilowatts of power to perform tasks that the human brain can do on a 20W power budget. One of the ways the brain achieves this efficiency is through event-based ``spiking''  information processing, which has inspired research into neuromorphic computing \cite{loihi2018,richter2024dynap,pehle2022brainscales,gonzalez2024spinnaker2}. 
However, for a long time there had been doubts about whether spiking neural networks (SNNs) can be trained by gradient descent, the gold standard in machine learning, due to the non-differentiable jumps of the membrane potential when spikes occur. 
Using approximations and simplifying assumptions and building up from single spikes and layers, gradient-based learning in SNNs has gradually been developed over the last 20 years, including the early SpikeProp algorithm~\cite{bohte2002} and its variants \cite{mckennoch2006,booij2005,xu2013supervised,xu2017online,mostafa2017supervised}, also applied to deeper networks \cite{lee2016training,wu2018spatio,sengupta2019going}, the Chronotron \cite{florian2012}, the (multispike) tempotron \cite{Gutig2006, Rubin2010, Gutig2016,Fil2020}, the Widrow-Hoff rule-based ReSuMe algorithm \cite{Ponulak2010,sporea2013supervised,zhang2017supervised} and PSD \cite{Yu2013}, as well as the SPAN algorithm \cite{mohemmed2012span,mohemmed2013training} and Slayer \cite{shrestha2018}.
Other approaches have tried to relate back-propagation to phenomenological learning rules such as STDP \cite{tanavei2019}, or to enable gradient descent by removing the abstraction of instantaneous spikes \cite{huh2018gradient}, or using probabilistic interpretations to obtain smooth gradients \cite{esser2015backpropagation}.
More recently, new algorithms in two main categories have been discovered. Many groups are proposing gradient descent-based learning rules that employ a surrogate gradient \cite{Zenke2018,Kaiser2020,Bellec2020} while others have developed novel ways of calculating exact gradients \cite{Wunderlich2021,Goeltz2020,Goeltz2021,Comsa2021}. 
The arrival of these new methods has made gradient-based learning a realistic prospect and could enable a transition to low-energy neuromorphic machine learning. 
In this paper we investigate scaling up Eventprop learning \cite{Wunderlich2021} to benchmark problems beyond the original proofs of concept. Eventprop leverages the adjoint method from optimisation theory, to calculate exact gradients in a backward pass that is -- like the forward pass -- a hybrid system of {\em per neuron} dynamical equations and discrete communication between neurons that only occurs at the times of sparse recorded spikes (Table \ref{table1}).

\begin{table}[h]
\begin{tabular}{lll}
  \toprule
  {\bf Free dynamics} & {\bf Transition } & {\bf Jumps at
    transition} \\
    & {\bf condition} & \\
  \midrule 
  Forward: \\
 (i) $\tau_{\text{mem}} \dot{V} = -V + I$ & $(V)_n - \vartheta = 0$,
   &
  $(V^+)_n = 0$ \\
  (ii) $\tau_{\text{syn}} \dot{I}= -I$ & $\big(\dot{V}\big)_n \neq 0$ & $I^+= I^- + W
  e_n$ \\
  \midrule
    Backward: \\
  (iii) $\tau_{\text{mem}} \lambda_V' = - \lambda_{V} - \frac{\partial l_V}{\partial V}$ & $t-t_k = 0$ &
  (v) $(\lambda_V^-)_{n(k)} = (\lambda_V^+)_{n(k)} +
  \frac{1}{\tau_{\text{mem}} (\dot{V}^-)_{n(k)}} \Big[
      \vartheta (\lambda_V^+)_{n(k)}$ \\
  (iv) $\tau_{\text{syn}} \lambda_I' = -\lambda_I + \lambda_V$ & & $\hphantom{(\lambda_V^-)_{n(k)} = } + \left(W^T
      (\lambda_V^+ - \lambda_I)\right)_{n(k)} + \frac{\partial
        l_p}{\partial t_k} + l_V^- - l_V^+ \Big]$ \\
  \midrule
    \multicolumn{3}{l}{{\bf Gradient of the loss:} 
    (vi) $\frac{d {\cal L}}{d w_{ji}} =-\tau_{\text{syn}} \sum_{t \in t_{\text{spike}}(i)} \lambda_{I,j} (t)$}  \\
  \bottomrule \\
\end{tabular} 
\caption{
Original Eventprop gradient calculation, adapted from \cite{Wunderlich2021} \label{table1}. $V$ and $I$ are the membrane potential and input current and $\lambda_V$ and $\lambda_I$ the corresponding adjoint variables. $\tau_{\text{mem}}$ and $\tau_{\text{syn}}$ are the membrane and synaptic time constants. $W$ is the weight matrix and $\vartheta$ the firing threshold. The dot denotes the derivative with respect to time and the prime the derivative backwards in time. Superscript ``-'' and ``+'' denote the values before and after a discontinuous jump. $l_p$ and $l_V$ are defined by the loss function, \cite{Wunderlich2021}, equation (1). }
\end{table}

The adjoint method also allows more flexibility than other exact-gradient approaches and generalizes to a wide class of event-based neural networks including event-based GRU models~\cite{subramoney2023}.
Besides using exact gradients rather than approximations, which some may find appealing, Eventprop also has attractive properties in terms of numerical efficiency, in particular for parallel computing: existing parallel algorithms for SNN simulation -- whose compute complexity scales predominantly with the number of neurons rather than the number of synapses -- can be employed for both forward and backward passes, and memory requirements only grow with the number of spikes rather than the number of timesteps in the trial.
An alternative approach to reducing the complexity of BPTT while still using surrogate gradients is to only propagate back gradients from `active' neurons~\cite{perez2021sparse,subramoney2023}.
However, as by definition, surrogate gradient functions are `wider' than the spikes they are a surrogate for, these approaches will never achieve the level of activity sparsity achieved by event-driven algorithms and thus have smaller memory and compute savings.

In this paper, we consider shallow SNNs of leaky integrate-and-fire (LIF) neurons and exponential synapses (Table \ref{table1}, ``Forward''), consisting of an input layer, one hidden layer and an output layer with non-spiking leaky integrator neurons. The weights of the synaptic connections in the networks are trained with the Eventprop algorithm \cite{Wunderlich2021}, see Table \ref{table1}.
We have implemented Eventprop in the GPU enhanced neural networks framework (GeNN) \cite{Yavuz2016,Knight2018} using the Python interface PyGeNN \cite{Knight2021} and, here, performed simulations using the CUDA backend for NVIDIA GPUs. Our code is available on Github \cite{genn_eventprop}.

We first reproduced the latency encoded MNIST \cite{LeCun1998a} classification task before moving on to the more challenging Spiking Heidelberg Digits (SHD) and Spiking Speech Commands (SSC) keyword recognition datasets \cite{Cramer2022}.
When working on the SHD dataset we noticed issues that arise from using the exact gradient for particular combinations of loss functions and task attributes. To overcome this, we extended Eventprop to a wider class of loss functions and, using this additional freedom, identified better-performing formulations of cross-entropy loss, including one that leads, in conjunction with augmentation, to state-of-the-art performance on SHD. We then successfully applied the same network to SSC.
Finally, we compared the time and memory requirements of training with Eventprop against training with BPTT, implemented using Spyx~\cite{heckel_spyx_2024} (currently the fastest framework for surrogate-gradient-based training).
We found that, when training SHD with $1$ ms timesteps, Eventprop was more than $3\times$ faster and used as little as $\sfrac{1}{4}$ of the GPU memory required by BPTT.

\noindent
Our main contributions in this work are:
\begin{itemize}
    \item We identified that loss functions that may otherwise be appropriate for a task may fail because of a ``spike deletion problem'' in SNNs of LIF neurons that are trained using exact gradients.
    \item We extended Eventprop to work with a wider class of loss functions including the form of ``sum'' and ``max'' loss widely used with BPTT.
    \item Within this wider class, we identified a variation of cross-entropy loss that enables successful Eventprop learning of SHD and SSC.
    \item We derived Eventprop gradient calculations for neuron membrane and synaptic timescales in addition to the known equations for weights.
    \item We investigated data augmentations, heterogeneous timescales and training timescales, as well as changes in network architecture to find a training setup that leads to state-of-the-art accuracy for LIF-based SNNs in SHD and SSC. 
    \item We developed an efficient GeNN implementation of Eventprop which exhibits good numerical scaling properties.
\end{itemize}
\vfill
\section{Results}
For the discussion that follows it is useful to recall how the Eventprop algorithm can be used to calculate the exact gradient of the loss function of a spiking neural network (SNN) with LIF neurons and exponential synapses. These networks are defined by a hybrid system of ordinary differential equations describing the dynamics of the state variables (the voltage $V$ and the current $I$), and discrete updates (``jumps'' or ``spikes'') to apply to the state variables when a certain threshold condition is met (see Table \ref{table1}, ``Forward'').
The Eventprop algorithm is derived from the adjoint method in optimization theory and can be applied to loss functions which either depend on spike times -- as LIF neuron's spike times depend in a differentiable way on weights -- or are the integral of a function of the voltage $V$ -- because then the integral can be used to smooth the non-existing derivative of the voltage at spike times.
Eventprop resembles BPTT in that it is a forward pass / backward pass algorithm but, in the backward pass, the derivative of the loss with respect to the weight parameters is calculated as a hybrid system of dynamic equations and discrete jumps of so-called adjoint variables (see Table~\ref{table1}, ``Backward''). 
There is one adjoint variable for each dynamic variable of the forward dynamics, i.e. $\lambda_V$ for the voltage $V$ and $\lambda_I$ for the current $I$. 
Intuitively, the values of the adjoint variables at the times $t_k$ of spikes in the preceding forward pass keep track of the loss that the transmission of that spike through a synapse had eventually caused i.e. the ``blame'' attributed to the spike and eventually the weight of the synapses it is transmitted by. 
For output neurons, this blame is directly added to $\lambda_V$ through the term $\frac{\partial l_V}{\partial V}$ and for all other neurons it is passed back from their post-synaptic partners through $W^T(\lambda_V^+-\lambda_I)$. 
The total assigned blame for each weight parameter $w_{ji}$ is then accumulated from the post-synaptic $\lambda_{I,j}$ over pre-synaptic spike times $t_i$, which makes intuitive sense as it was at those times that $w_{ji}$ contributed to the activity of the post-synaptic neuron and hence the eventual loss in the trial.
The dynamics equations of the adjoint variables track how $I$ has caused deflections in $V$ and hence loss -- either through causing spikes or directly in the loss function in the case of output neurons.

It is straightforward to implement this algorithm in a spiking neural network simulation framework because the backward pass has the same computational structure as the forward pass.
We have implemented the Eventprop algorithm in the GeNN framework 
and first reproduced the results of Wunderlich and Pehle \cite{Wunderlich2021} on the latency-encoded MNIST \cite{LeCun1998a} dataset. In this benchmark, the $28\times 28$ grayscale images are unrolled into a 1D input vector and the grayscale of each entry is translated linearly into the spike time (``latency'') of one input neuron.
Accordingly, each of the $784$ input neurons spikes only once (see Figure S1). We used the average cross-entropy loss 
\begin{eqnarray}
    {\cal L}_{\text{x-entropy}} = -\frac{1}{N_{\text{batch}}} \sum_{m=1}^{N_{\text{batch}}} \int_0^T \log \left( \frac{\exp\left(V_{l(m)}^m(t)\right)}{\sum_{k=1}^{N_{\text{class}}} \exp\left(V_{k}^m(t) \right)} \right) \, dt, \label{xentropyloss}
\end{eqnarray}
where $N_{\text{batch}}$ is mini-batch size, $m$ the trial index, $T$ the trial duration, $V^m_*$ the output voltage of output $*$ in the $m^\text{th}$ trial, and $l(m)$ the correct class label. 
This is appropriate for Eventprop as it is of the form ${\cal L}_{\text{x-entropy}} = \int_0^T l_V(V) dt$ with 
\begin{eqnarray}
    l_V(V) = -\frac{1}{N_{\text{batch}}} \sum_{m=1}^{N_{\text{batch}}} \log \left( \frac{\exp\left(V_{l(m)}^m(t)\right)}{\sum_{k=1}^{N_{\text{class}}} \exp\left(V_{k}^m(t) \right)} \right). 
\end{eqnarray}
We used a three-layer feedforward LIF network (784 -- 128 -- 10 neurons) and achieved a similar classification performance on the test set ($97.8 \pm 0.1$\% correct -- mean $\pm$ standard deviation in $n=10$ repeated runs) as in \cite{Wunderlich2021} ($97.6 \pm 0.1$\% correct). This independently reproduces their work and demonstrates that our discrete-time implementation with $1$ ms timesteps is precise enough to achieve the same performance as their event-based simulations in this task.

We then considered SHD. We again used ${\cal L}_{\text{x-entropy}}$ to train a variety of three-layer networks with differently sized hidden layers with and without recurrent connectivity, and with a variety of meta-parameter values.
However, the trained networks only performed close to chance level (e.g. training performance $10.9\pm 1.3$\% correct ($n=10$) after 200 epochs for a feedforward network with 256 hidden neurons. Chance level is $5$\%). To understand this failure, we inspected the learning dynamics of the network in more detail. 

\begin{figure}
    \includegraphics[width=\textwidth]{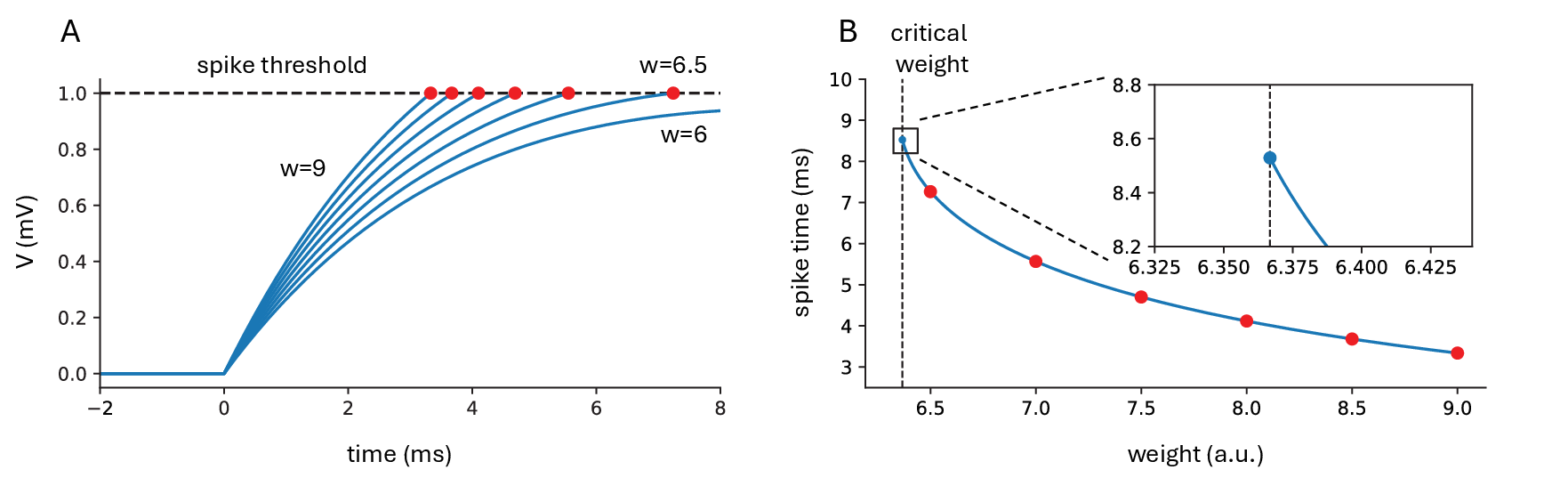}
    \caption{Relationship of spike times in LIF neurons and weights of incoming synapses. \textbf{(A)} voltage $V$ of a LIF neuron as a function of time in response to a single incoming spike through a synapse of weight $w$. The higher $w$ the earlier the spike threshold is crossed and a spike is emitted (red dots). For $w=6$ the threshold is never crossed. \textbf{(B)} Time of threshold crossing as a function of the incoming weight in the scenario shown in A. The red dots match those in panel A. With decreasing $w$ the spike time increases continuously but then stops abruptly at a critical weight value when the spike threshold can't be reached any more. Crucially, the slope of the curve is finite before this point (see inset) so that there is no indication in the gradient about the existence of the critical point. \label{fig:LIF_spiking}}
\end{figure}

We discovered that the failure was caused by a fundamental property of gradient descent with exact gradients in SNNs of LIF neurons.
The jump conditions in the hybrid system defining the neurons (see Table \ref{table1} ``Forward'') are typically expressed as threshold crossings of the voltage variable $V$. 
The time of a threshold crossing -- and hence when jumps (spikes) occur -- is a continuous function of the weights of incoming synapses to the neuron (Figure \ref{fig:LIF_spiking}). However, as can be seen in Figure \ref{fig:LIF_spiking}B, when decreasing weights for LIF neurons, the threshold crossing will stop suddenly at a ``critical value'' of incoming weights.
The gradient of the spike time with respect to the weights in the vicinity of this critical value is finite (Figure \ref{fig:LIF_spiking}B inset) and so does not contain any indication about the existence of the critical value nearby. Therefore, if a gradient descent procedure follows the gradient in finite steps, it will inadvertently cross the critical value if the gradient points to lower weight values. This will erase the spike. Equally, when weights are increased, previously non-existing spikes will be created regardless of whether they increase or decrease loss.
This obviously poses a potentially serious problem because descending the loss gradient only reduces loss if spike times change in accordance with the continuous function of the incoming weights that the gradient is based on. 
If the critical value is crossed instead, the erasure or creation of a spike will have unpredicted effects and might increase the loss.

\begin{figure}
\includegraphics[width=\textwidth]{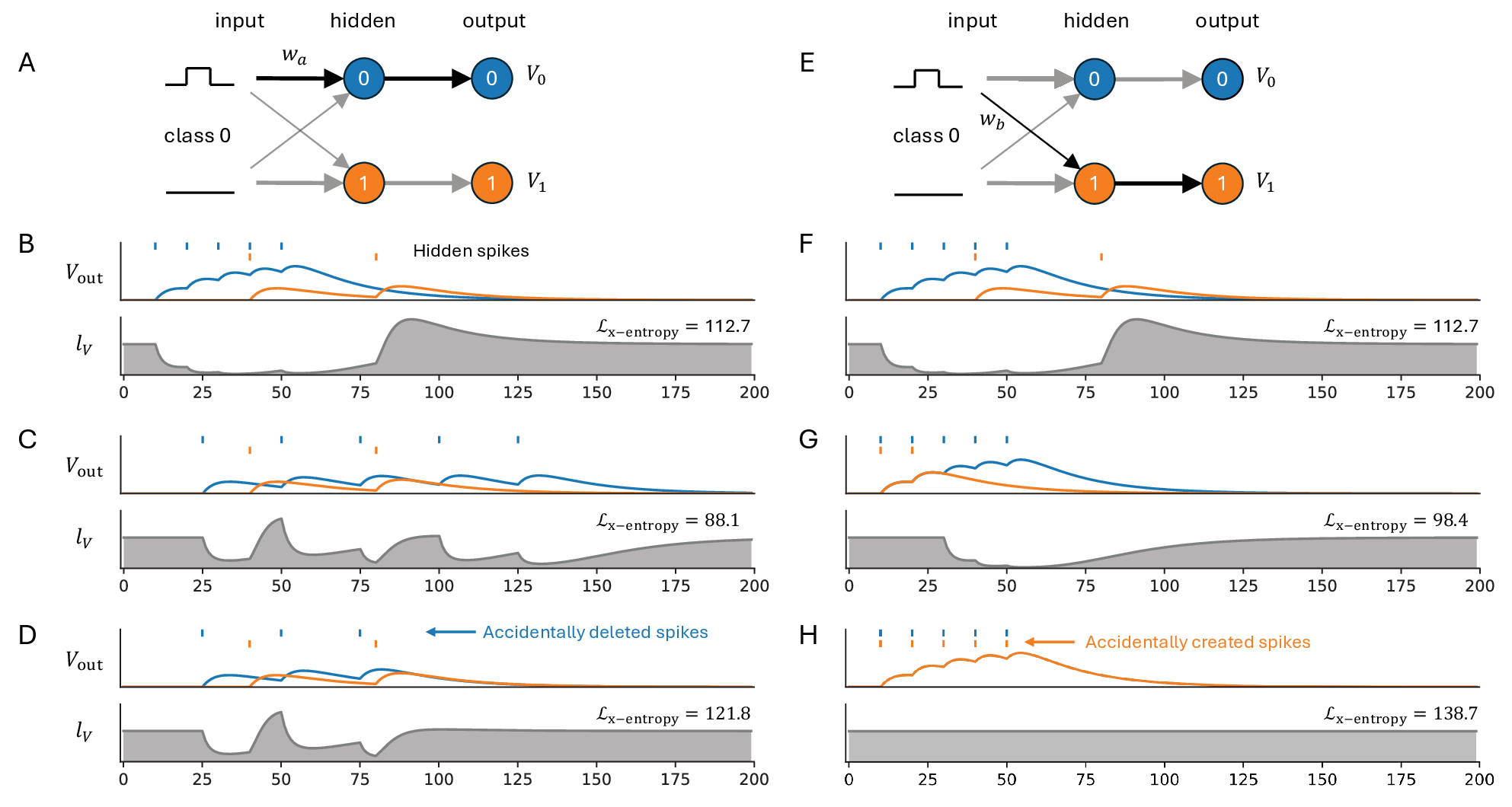}
\caption{Gedankenexperiment illustrating the problem of accidental gradient ascent. \textbf{(A)} minimal network for a two-class classification problem with one hidden layer. \textbf{(B)}~original assumed scenario of spikes in hidden neurons, the resulting output voltages $V_0, V_1$ and the corresponding loss term $l_V(V_0,V_1)$. \textbf{(C)}~fictitious scenario of spread-out spikes in hidden neuron $0$ that would have a lower loss. \textbf{(D)}~actual outcome of spreading out spikes by lowering $w_a$ with accidentally deleted spikes and hence higher loss than at the beginning. \textbf{(E)-(H)}like (A)-(D) but considering the spikes in hidden neuron $1$ where the gradient points to moving spikes closer/ forward in time in hidden neuron $1$,  leading to increases in $w_b$ and hence detrimental accidental creation of spikes. \label{fig:toy_example}}
\end{figure}

We have collected evidence that this fundamental issue caused the learning failure of SHD with average cross-entropy loss. In essence, the averaged cross-entropy loss is optimised with respect to the spike times of hidden neurons when 
spikes of hidden neurons that have positive weight to the correct output neuron are spread out in time. This is because the cross-entropy expression has diminishing returns for multiple spikes occurring at the same time. But, spreading out hidden spikes by increasing and decreasing weights can lead to their deletion as explained above, which inadvertently increases the loss. How this can play out is illustrated in the Gedankenexperiment of Figure \ref{fig:toy_example}. The task in this toy example is to distinguish two classes of inputs, class $0$, where input $0$ emits a constant current pulse for about a third of the trial and input $1$ emits zero and class $1$ where the roles are reversed.
Output $0$ should be most activated by class $0$ and output $1$ by class $1$.
We assume that the network has already learned the obvious solution to activate hidden neuron $0$ for class $0$ and this hidden neuron strongly activates output neuron $0$ (see Figure  \ref{fig:toy_example}A).
If we now consider the cross-entropy loss as illustrated in Figure \ref{fig:toy_example}B, it is clear that the loss can be further decreased by spreading out the spikes of hidden neuron $0$ to increase the output $V_0$ throughout more of the trial, as shown in Figure \ref{fig:toy_example}C.
The gradient hence points to weakening $w_a$ to delay spikes in hidden neuron $0$.
This means that eventually, spikes get deleted and we end up with a {\em higher loss} as shown in Figure \ref{fig:toy_example}D. Note how the presumed loss improvements for spread-out spikes in Figure \ref{fig:toy_example}C over Figure \ref{fig:toy_example}B relate to the saturating nature of cross-entropy terms and hence diminishing returns for closely packed additional spikes in hidden neuron $0$.
Similarly, spikes in hidden neuron $1$ would be moved together/forward to decrease loss (Figures \ref{fig:toy_example}F,G) but this will increase $w_b$ and generate additional spikes, so that in the end the loss is {\em larger than before} (Figure \ref{fig:toy_example}H).
In essence, the exact gradient of the loss is such that descending it inadvertently crosses the critical value for spike deletion and spike creation and {\em increases loss} rather than decreasing it.

\begin{figure}
    \centering
    \includegraphics[width=\textwidth]{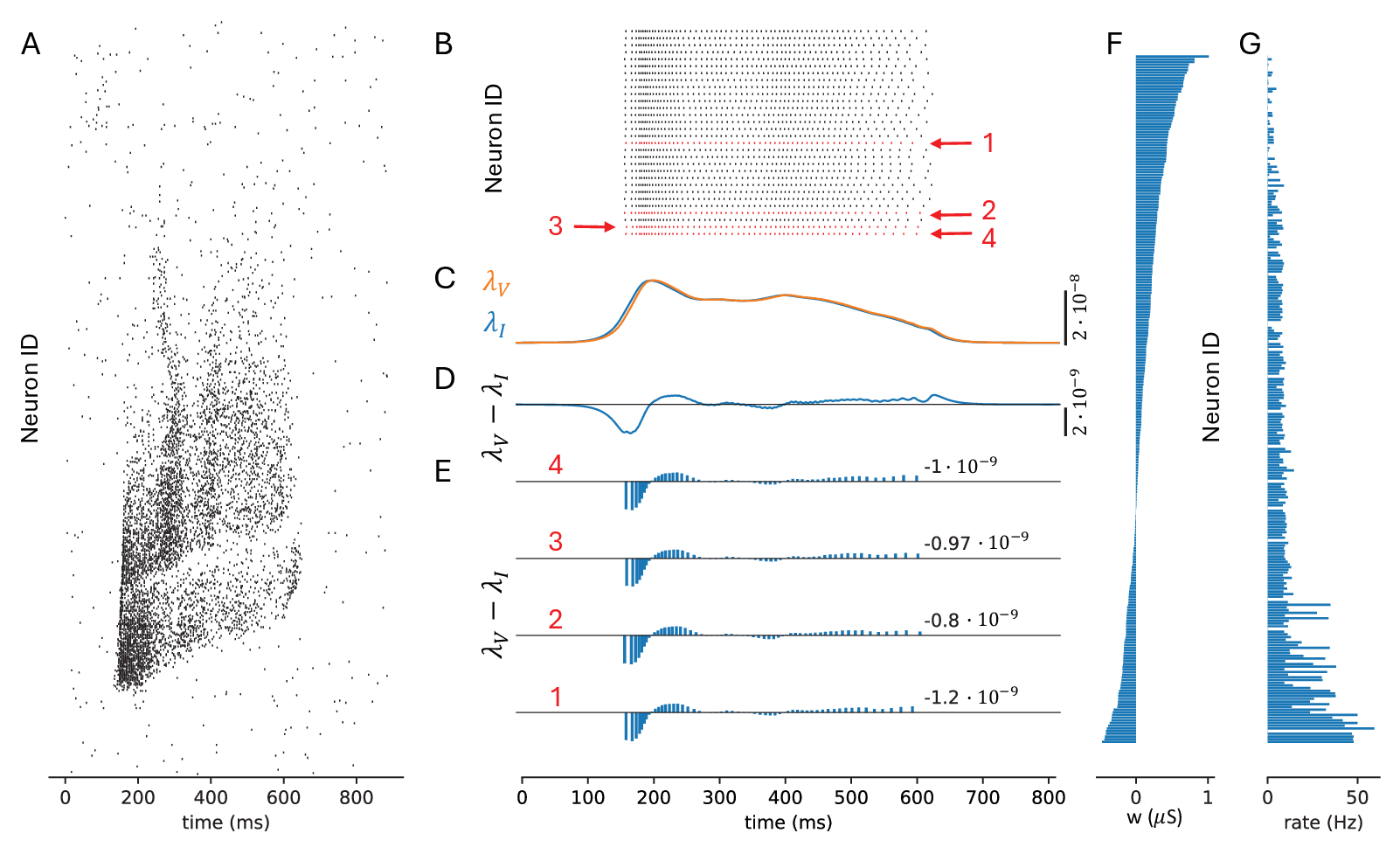}
    \caption{Illustration of the mechanism that leads to unhelpful spike deletions in hidden neurons. \textbf{(A)} Spike raster of a typical input pattern of class 0. \textbf{(B)} Spike raster of the hidden layer in response to an input of class 0 (showing a subset of $30$ of $256$ neurons for better visibility). Red highlighted neurons are those that are most active on average for class 0 inputs and correspond to the panels shown in E. \textbf{(C)} $\lambda_V$ (orange), $\lambda_I$ (blue) of output neuron $0$ in the corresponding backwards pass plotted against forward time, i.e. integration proceeds from the right to the left. During backwards integration, $\lambda_V$ increases rapidly from 0 to the value corresponding to all output voltages being $0$ and $\lambda_I$ follows $\lambda_V$ (around t=1400, not shown). When the stored spikes are encountered, $\lambda_V$, $\lambda_I$ increase further as the model is not yet trained and the correct output voltage does not dominate in the response. \textbf{(D)} The difference $\lambda_V-\lambda_I$ of output $0$ that is transported to the neurons in the hidden layer. \textbf{(E)} $\lambda_V-\lambda_I$ values arriving at the four most active neurons (marked in red in B) when transported during a stored spike, shown as bars. The numbers indicate the sum of all bars, which relates to the direction of the total change in excitation the hidden neurons receive. All values are negative, i.e. neurons with positive weights towards the correct output $0$ will become less activated for this and similar inputs of class $0$ after the learning update and hidden neurons with negative weight will become more active -- exactly opposite to what one would expect for efficient learning. \textbf{(F)} distributions of weights from hidden neurons onto neuron $0$ after $30$ epochs of training on class $0$. \textbf{(G)} Average firing rate of hidden neurons, in response to inputs of class $0$ during the last mini-batch of the same training. Neurons are in the same order in F and G (sorted by their weight onto output $0$).}
    \label{fig:axe1}
\end{figure}

In the Eventprop algorithm, the updates of weights to the hidden neurons that eventually lead to the spike deletion problem are driven by the updates $w\cdot (\lambda_V - \lambda_I)$.
From table \ref{table1} equations (v),(iii),(iv) and (vi), it is clear that hidden neurons that have a positive weight towards the correct output neuron (the case of hidden neuron 0 above) will receive less excitation after learning updates when $\lambda_{V}-\lambda_{I}$ is on average less than $0$ in a trial, and those with a negative weight to the correct neuron or a positive weight to other output neurons (hidden neuron 1 above) will receive more. 
Figure \ref{fig:axe1}A-E shows an example of a typical trial of our network trained with ${\cal L}_{\text{x-entropy}}$ on SHD in the early learning phase. In this example, an input of class $0$ was shown but output $0$ was not the highest. Hence, $\lambda_{V,0}$, and hence $\lambda_{I,0}$ increase in the backward pass during the period where input and hidden spikes had occurred. This leads to transported error signals $\lambda_{V,0}-\lambda_{I,0}$ that {\em are on average less than $0$} (Figure \ref{fig:axe1}D, E and therefore decrease the drive of hidden neurons with a positive weight to the correct output $0$ and increase the drive of neurons with negative weights to output $0$. In other words, the hidden neurons that are driving the correct output are gradually switched off and those that suppress the correct output are switched on -- exactly as in the Gedankenexperiment above and the opposite of what is needed. Close inspection of Figure\ref{fig:axe1}E illustrates that the negative error signal stems from the sharp drop of $\lambda_V$ in backwards time, related to the sudden onset of spikes at the beginning of the trial and the preceding silent period. Similar effects occur for correct trials and the trailing silent period.

To investigate this effect more explicitly we trained the network with input class $0$ only and inspected the activity of hidden neurons and their weights towards the correct output $0$. The number of spikes in the hidden neurons after training (Figure \ref{fig:axe1}G and their output weights onto output $0$ (Figure \ref{fig:axe1}F) are strongly negatively correlated, Pearson correlation coefficient $-0.707$. The highest output weights are from hidden neurons that no longer respond to the class of inputs that the output neuron is supposed to represent -- they have been switched off due to the gradient descent in the hidden layer.

The observed learning failure is hence caused by the combination of an absence of information about the creation or deletion of spikes in the exact gradient and the structure of the loss function that drives the network towards the critical weight values where useful spikes are deleted essentially because the loss function aims to minimise cross-entropy at \emph{all} times during the trial, even during the silent periods at the start and end. In support of this hypothesis, if we remove the silent period at the end of the trial by training the network on the first $400$ ms of each SHD digit, we see a somewhat better training performance ($30.2\pm 1.2$\% correct ($n=10$)).

To avoid the problems incurred with ${\cal L}_{\text{x-entropy}}$ altogether we need to remove the arguably unnecessary goal of being able to make a correct classification at all times during the trial, particularly where this is impossible due to periods of silence in the input neurons.
A natural loss function to consider with this goal in mind is the cross-entropy of the sum or integral of the membrane potentials of non-spiking output neurons, which has been used widely with BPTT \cite{Zenke2021},
\begin{eqnarray}
    {\mathcal L_{\text{sum}}} = - \frac{1}{N_{\text{batch}}} \sum_{m=1}^{N_{\text{batch}}} \log \left( \frac{\exp\left(\int_0^T V_{l(m)}^m(t) dt\right)}{\sum_{k=1}^{N_{\text{out}}} \exp\left(\int_0^T V_{k}^m(t) dt\right)} \right), \label{eqn:sumloss}
\end{eqnarray}
where $V_k^m$ is the membrane potential of output $k$ in trial $m$ and $T$ trial duration.
With ${\cal L}_{\text{sum}}$, the contribution of each spike has (almost) no dependency on when it occurs because the contribution of each spike to the output voltages sums up linearly without saturation. There is hence (almost) no pressure for changes to hidden neurons' input weights, avoiding the learning failure. However, ${\cal L}_{\text{sum}}$ is not supported in the normal Eventprop framework. This prompted us to extend the Eventprop framework to a wider class of class functions.

\subsection{Additional loss functions in Eventprop}

We extend Eventprop to losses of the shape
\begin{eqnarray}
    {\cal L}_F = F\left(\textstyle \int_0^T l_V(V(t),t) \, dt\right), \label{eqn:loss_gen}
\end{eqnarray}
where $F$ is differentiable and $l_V$ can be vector-valued, e.g. $l_V = V$ as in the loss functions used below. At first, it seems impossible to ``swap'' the exponential function and integration to arrive at an Eventprop-compatible loss of the form $\int_0^T l_V(V) dt$. However, using the chain rule and re-organising terms in an appropriate way (see Appendix A for a detailed derivation) we can derive an extended scheme that is identical to the original Eventprop method except that equation (iii) in table \ref{table1} is replaced by 
\begin{eqnarray}
\tau_{\text{mem}} \lambda_{V,j}' &=& - \lambda_{V,j} - \frac{\partial F}{\partial (\int l_V dt)} \cdot \frac{\partial l_V}{\partial V_j}. \label{eqn:newdyn1} 
\end{eqnarray}
Equipped with the extended scheme, we implemented loss functions for classifying based on the sum of integrated membrane voltage of non-spiking output neurons (\ref{eqn:sumloss}) or on the maximum, ${\cal L}_{\text{max}}$, as in \cite{Wunderlich2021}, equation (4). 
We also compare against using spiking output neurons and classifying depending on the first spike with the ${\cal L}_{\text{time}}$ loss from \cite{Wunderlich2021}, equation (3).
We found that all loss functions performed reasonably well for the latency MNIST task, with a slightly lower performance when using ${\cal L}_{\text{time}}$ (Figure S1).

\begin{figure}
    \centering
    \includegraphics[width=\textwidth]{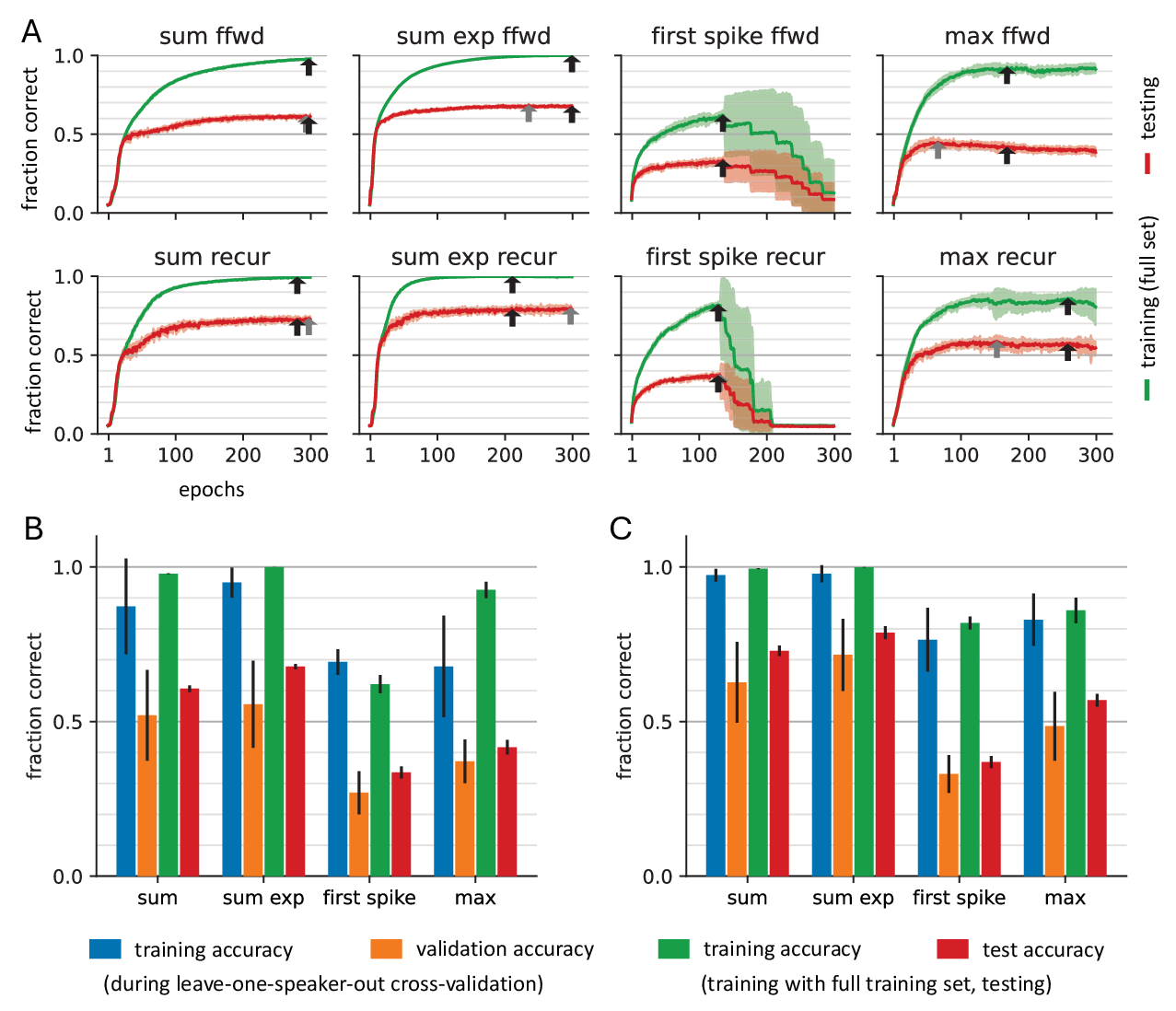}
    \caption{Summary of initial SHD classification results with a simple network, including regularisation only. \textbf{(A)} Learning curves for training (green) and testing accuracy (red). "ffwd" are feed-forward networks, "recur" recurrent networks. Curves are the mean of $8$ repeated runs with different random seeds and shaded areas indicate one standard deviation around the mean. The black arrows indicate the location of the best-achieved training accuracy and point to the values summarised in panels B and C. The grey arrows marks the highest test accuracy. \textbf{(B)} average accuracies in feedforward networks at the epoch with the best validation error for cross-validation and at the epoch with the best training error for train/test (black arrows in A). Values are the average across 10 folds in leave-one-speaker-out cross-validation and the average across 8 independent runs for train/test. Error bars are the corresponding standard deviations. \textbf{(C)} as B but for recurrent networks. The results for the failing ${\cal L}_{\text{x-entropy}}$ loss were omitted in this figure to avoid too much clutter. \label{fig:SHD}}
\end{figure}

\subsection{Spiking Heidelberg Digits}
We then returned to the SHD benchmark. We optimised the meta-parameters of the models with each of the loss functions using grid searches in a 10-fold cross-validation approach: In each fold, we trained the network on $9$ of the speakers and tested it on the examples spoken by the $10$th speaker (see tables \ref{table2}, \ref{table3}). Then, we measured training and testing performance with the full training- and test set. The results are shown in Figure \ref{fig:SHD}.

For all loss functions the recurrent networks performed better than the feedforward networks (Figure \ref{fig:SHD}C versus B), in line with observations in networks trained with BPTT \cite{Zenke2021}.
In both types of networks, apart from the completely failing ${\cal L}_{\text{x-entropy}}$ loss, the worst performance and least reliability were observed for ${\cal L}_{\text{time}}$, followed by ${\cal L}_{\text{max}}$. ${\cal L}_{\text{sum}}$ performed competitively with respect to the results reported in the literature \cite{Zenke2021} and the results with the Realtime Recurrent Learning~(RTRL)-approximation e-prop \cite{Bellec2020}, obtained in our lab \cite{Knight2022}, especially when used with recurrent connectivity. However, the performance was not quite as good as the competitors and, when we inspected the learning dynamics, we observed that learning for ${\cal L}_{\text{sum}}$ is comparatively slow despite an increased learning rate $\lambda = 2\cdot10^{-3}$ (see table \ref{table3}). The optimised regularisation strength $k_{\text{reg}}$ (see Methods) of the hidden layer is also orders of magnitudes smaller in this model than in others. Both observations indicate that the gradients flowing from the output towards the hidden layer are very small. On reflection, this effect can be easily understood when realising that, due to the definition of ${\cal L}_{\text{sum}}$, the timing of hidden spikes has (almost) no effect on their contribution to the overall loss. Every post-synaptic potential (PSP) causes the same added (or subtracted for negative synaptic weight) area under the membrane potential and hence the same increase or reduction in loss.
The only PSPs for which this is not the case are those that are `cut off' at trial end.
By moving \emph{these} PSPs forwards or backwards in time, the amount cut off their area can be changed, resulting in tiny contributions to the gradient.

Based on this insight, we improved the gradient flow to the hidden layer by adding a weighting term to ${\cal L}_{\text{sum}}$ that would make earlier PSPs more effective for increasing or reducing loss. We tried four different weightings, linearly decreasing, exponentially decreasing, sigmoid and proportional to the number of input spikes at each timestep. In preliminary numerical experiments, we found that exponential weighting performed well (data not shown),
\begin{eqnarray}
    {\mathcal L_{\text{sum\_exp}}} = - \frac{1}{N_{\text{batch}}} \sum_{m=1}^{N_{\text{batch}}} \log \left( \frac{\exp\left(\int_0^T e^{-t/T} V_{l(m)}^m(t) dt\right)}{\sum_{k=1}^{N_{\text{out}}} \exp\left(\int_0^T e^{-t/T} V_{k}^m(t) dt\right)} \right). \label{eqn:sumexploss}
\end{eqnarray}

As shown in Figure \ref{fig:SHD}, this leads to accuracies ($67.8\pm 0.9\%$ feedforward, $78.8\pm 2.0\%$ recurrent) slightly lower than previous SNN results with BPTT \cite{Zenke2021} and comparable to our own results with e-prop \cite{Knight2022}.

\subsection{Applying machine learning tools for better accuracy}
Once we overcame the major obstacle of finding an appropriate loss function, we applied tools from the machine learning toolbox to maximise classification accuracy. First, we applied four different types of data augmentation (see methods for details): Global random shifts across input channels, randomly re-assigning individual spikes to nearby input channels, compressing or dilating the duration of stimuli and generating new input patterns by blending together two random examples of the same class. Only ``shift'' and ``blend'' augmentations proved effective.

Next, we investigated different network structures, including different hidden layer sizes, multiple hidden layers and a ``delay line'' where $10$ copies are made of each input neuron and copy $n$ is activated with a delay of $n\cdot 30$ ms. All networks had fully recurrent hidden layers based on the observations above. We found in initial scans that multiple layers did not improve accuracy (data not shown) and did not pursue this further.

We also investigated the role of the simulation timestep, noting that competitors are using timesteps as large as $10$ ms \cite{hammouamri2023learning} or even $25$ ms \cite{bittar2022surrogate}. We found that $1$ms and $2$ms timesteps performed best, with acceptable results for $5$ ms steps but clear degradation for $10$ ms (see figure S2).

Finally, we derived equations for training timescales $\tau_{\text{mem}}$ and $\tau_{\text{syn}}$ (see appendix~\ref{sec:taulearn}) and investigated homogeneous and heterogeneous initialisation for timescale parameters following \cite{perez2021neural}.

As a result of our extensive experimentation, we achieved the maximum cross-validation accuracy of $92$\% and, for the same parameters, $93.5\pm 0.7$\% test accuracy ($n=8$). 

\begin{figure}
    \includegraphics[width=\textwidth]{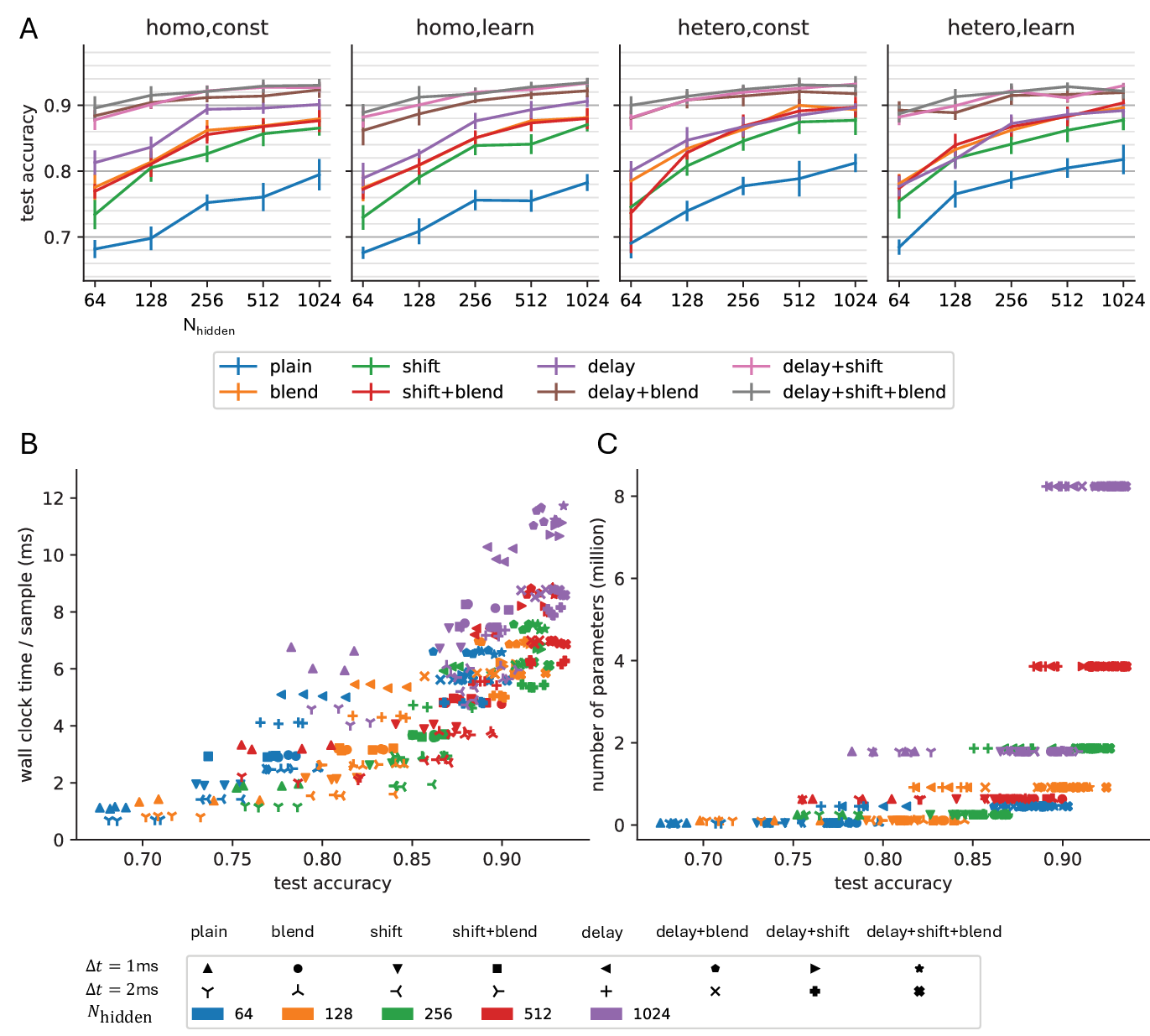}
    \caption{Ablation study on the SHD dataset. \textbf{(A)} accuracy on the test set as mean (line) and standard deviation (errorbars) of 8 independent runs with different random number seeds. The panels are for different combinations of homogeneous and heterogeneous initialisation of $\tau_{\text{mem}}$ and $\tau_{\text{syn}}$ and for static or trained $\tau$ values as indicated. The different coloured lines correspond to the different augmentations applied as shown. \textbf{(B)} Wall clock time per sample during training as a function of test accuracy for all the different conditions as indicated by the symbols and colours. This data includes runs with $\Delta t = 1$ms and $\Delta t= 2$ms. \textbf{(C)} Number of parameters, including tau values where trained, of the different networks as a function of the final test accuracy. Both B and C use the mean accuracy over 8 independent runs as in A.}
    \label{fig:final_SHD}
\end{figure}

We then investigated the contributions of the different mechanisms to classification accuracy in an ablation experiment. We started from the best solution and scanned the following parameters in all combinations: blend augmentation, shift augmentation, delay line, initialisation of $\tau$ and training $\tau$, using a rigorous validation methodology (see methods).
Figure \ref{fig:final_SHD}A shows the results. Without data augmentations or input delay line (Fig \ref{fig:final_SHD}A ``plain'', blue), we barely surpass $80$\% accuracy and the size of the hidden layer is important for better performance. Adding heterogeneous $\tau$ values and doing so in conjunction with $\tau$ learning improves the results in agreement with earlier results \cite{perez2021neural}. Adding the augmentations, the networks fall into two clusters of medium performance networks (``blend'',``shift'',``shift+blend'',``delay'') and best performance networks (``delay+blend'',``delay+shift'',``delay+shift+blend''). In this group of best performers, the hidden layer size matters less -- even a network with only $64$ hidden units can achieve almost $90$\% test accuracy -- and, interestingly, the advantages of heterogeneous $\tau$ and $\tau$ learning completely disappear. As Perez et al.~\cite{perez2021neural} were operating in the lower accuracy regime of our ``plain'' network, this is consistent with their results but casts a different light on the overall assessment of the importance of heterogeneity and timescale learning.

We then analysed the time taken to train the different networks in comparison to test accuracy (Figure \ref{fig:final_SHD}B). Unsurprisingly, there is a clear positive correlation between high accuracy and high runtime. Interestingly, however, there are networks with considerably lower than maximum runtime and yet almost the best accuracy (512 hidden neurons, ``delay+shift'' and ``delay+shift+blend'' -- fat crosses and plusses in red). 

When comparing test accuracy to the number of trained parameters (Figure \ref{fig:final_SHD}C), clearly more parameters typically lead to better results, though this levels out for networks larger than $512$ hidden neurons. Even some of the networks with $128$ hidden neurons and less than $1$ million parameters are still competitive.

\subsection{Spiking Speech Commands}
To test the generality of our observations, we also classified the words in the spiking speech commands (SSC) dataset~\cite{Cramer2022}. The SSC dataset consists of the Google speech commands data \cite{warden2018speech}, encoded using the same cochlea model used for SHD.

In preliminary experiments using ``plain'' networks, we observed that, unlike SHD, the SSC dataset is prone to underfitting. This indicated that these networks were too small to fully capture this data. However, after adding the additional delay-line input described above, more typical over-fitting was observed and we again addressed this with augmentations.

For SSC we obtained the best validation performance (early stopping) of $74.7$\% and for the parameter set in question $74.1\pm 0.9$\% test accuracy as measured in $8$ independent runs.

\begin{figure}
    \centering
    \includegraphics[width=\textwidth]{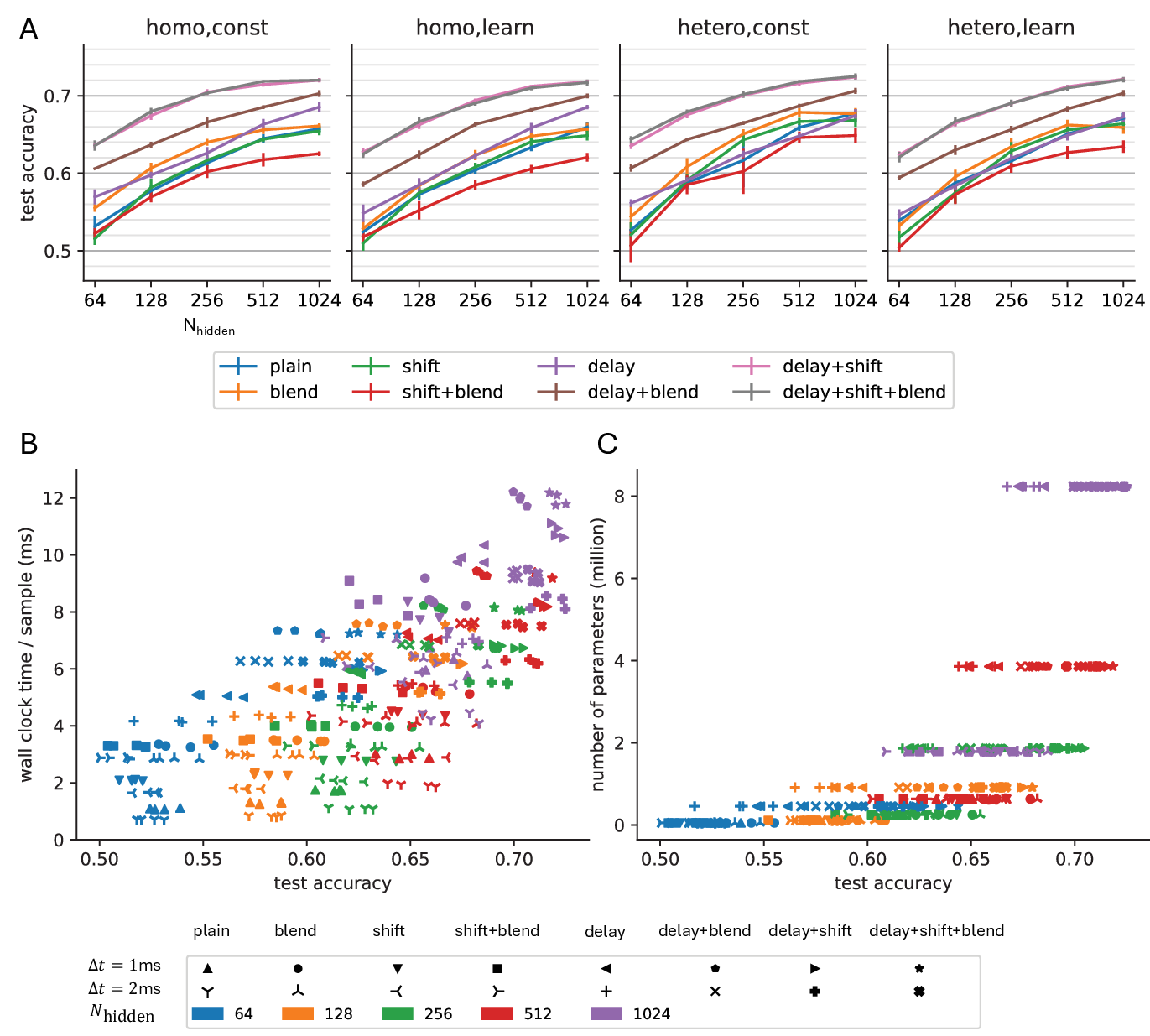}
    \caption{Ablation study on the SSC dataset. \textbf{(A)} accuracy of classification, \textbf{(B)} wall clock time versus accuracy, \textbf{(C)} number of parameters versus accuracy, all as in Figure \ref{fig:final_SHD}}.
    \label{fig:final_SSC}
\end{figure}

We then again performed a rigorous ablation study (figure \ref{fig:final_SSC}A to analyse the contributions of the different augmentations). Here, heterogeneous and learnt $\tau_\text{mem}$ and $\tau_\text{syn}$ do not appear to have much effect. The networks get better with more augmentations, except for the ``blend'' augmentation which does not seem to be very effective on SSC and even reduces test accuracy when used on its own (red lines versus blue lines in figure \ref{fig:final_SSC}A. This may be because SSC is obtained from a much larger number of speakers so adding additional examples obtained by ``blend'' is both less important and -- if blended examples are made from very different examples -- could produce unhelpful out-of-distribution examples.
Another difference to the observations on SHD is that the hidden layer size remains important, even for the best-performing networks, likely reflecting that SSC is a harder dataset. 
We also note that there is an even stronger separation than for SHD with ``delay+shift'' and ``delay+shift+blend'' clearly achieving significantly higher accuracy than any of the other networks.

When analysing runtime as a function of test accuracy (figure \ref{fig:final_SSC}B), we see the same trends as for SHD, though the cloud of data points is broader, reflecting the more pronounced differences in test accuracy between networks. Also, for SSC, the largest networks (purple) are more clearly achieving the best accuracy but there are networks that work equally well with $\Delta t = 2$ ms (purple fat plusses) as with $\Delta t = 1$ ms which offers some reduction in computational costs. When comparing the number of parameters and test accuracy (Figure \ref{fig:final_SSC}C), the relationship is clear with larger numbers of parameters leading to higher accuracy although it is worth noting that, between models with the same number of parameters, there are marked differences in the accuracy they achieve.

\subsection{Benchmarking against back-propagation through time and e-prop}
Back-propagation through time (BPTT) using surrogate gradients is the currently most popular approach to training SNNs with gradient descent. Another common approach is to use an approximation of RTRL such as e-prop~\cite{Bellec2020}. We compare the memory and time requirements of training models on SHD using Eventprop in GeNN against Spyx~\cite{heckel_spyx_2024} -- which currently has the fastest implementation of BPTT -- and our own implementation of e-prop -- which we have previously shown to be competitive against older, PyTorch-based BPTT libraries~\cite{Knight2022}.

\begin{figure}
    \centering
    \includegraphics[width=\textwidth]{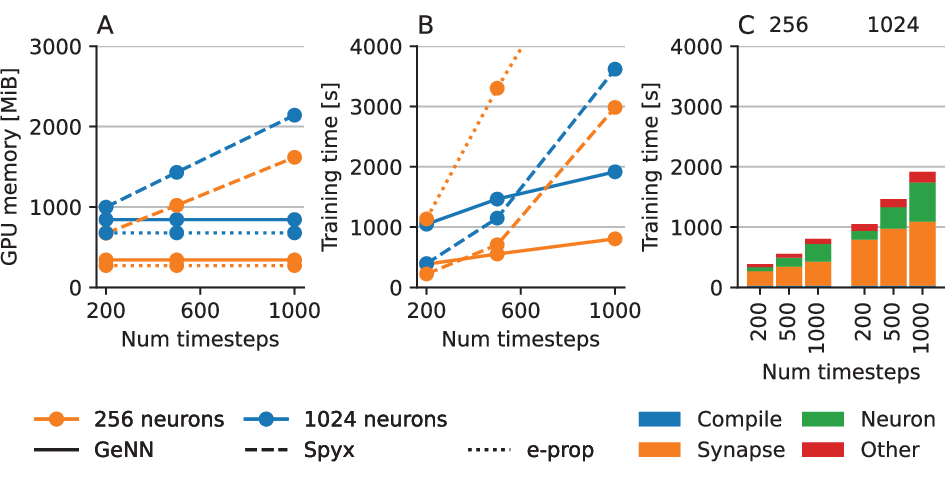}
    \caption{Comparing SHD training cost of EventProp implemented in GeNN against BPTT in Spyx and e-prop, also implemented using GeNN. \textbf{(A)} Peak GPU memory usage, \textbf{(B)} Time to train 100 epochs, \textbf{(C)} Scaling of different components of GeNN simulation. All experiments were performed on a workstation with NVIDIA RTX A5000 GPU. All models use batch size 32.}
    \label{fig:spyx_genn}
\end{figure}

Figure~\ref{fig:spyx_genn}A shows how GPU memory usage scales with the number of timesteps for two sizes of ``plain'' networks.
Because the memory usage of the EventProp backward pass scales with the number of events and this remains roughly constant, irrespective of the number of timesteps, the EventProp memory usage remains essentially constant.
However, the memory complexity of the Spyx simulation grows linearly with the number of timesteps as expected for BPTT~\cite{Zenke2021b}.
Furthermore, it is worth noting that, in all of our experiments, we allocate enough memory for 750 events per hidden neuron per trial which is conservative, if not excessive, and, when we are measuring memory, we are running JAX using the very slow ``platform'' memory allocator which uses significantly less memory than the faster strategy typically used.
Because e-prop does not require a backward pass, it requires slightly less memory than EventProp but this saving is relatively minor.

Figures~\ref{fig:spyx_genn}B and C show how -- for the same networks -- training time scales with the number of timesteps.
When using GeNN's hybrid simulation scheme, the time spent on event-based synaptic processing only grows very slowly with the number of timesteps (orange in Figure~\ref{fig:spyx_genn}C) and the cost of the time-driven neuron updates grows linearly with the number of timesteps (green). Overall, this results in sub-linear scaling with time.
In contrast, due to the nature of BPTT, training time in Spyx scales super-linearly with the number of timesteps. 
While e-prop only scales linearly with the number of timesteps, it does so very rapidly as the complexity of each timestep depends quadratically on the number of neurons (Figure~\ref{fig:spyx_genn}B). 
Finally, whereas the compilation time in GeNN is entirely independent of the number of timesteps, JAX's Just-In-Time compilation process appears to scale poorly with increasing numbers of timesteps.

\section{Discussion}
In this article, we have presented learning results for the Eventprop algorithm implemented in GeNN. We have identified issues when training spiking neural networks using the exact gradient calculated with Eventprop and the average cross-entropy loss function ${\cal L}_{\text{x-entropy}}$ on the SHD dataset. As explained in detail in the Results, this loss function led to the silencing of the most relevant hidden neurons. We overcame the problems by deriving an extended Eventprop algorithm that allows more general loss functions and found that the cross-entropy loss of average output voltages ${\cal L}_{\text{sum}}$ typically used with BPTT allowed successful learning but learning was slow and somewhat unreliable. We identified that this was due to an almost complete lack of gradient flow towards the hidden layer, so went full circle and augmented ${\cal L}_{\text{sum}}$ with a beneficial weighting term  (resulting in ${\cal L}_{\text{sum\_exp}}$) and \emph{finally} observed fast and reliable learning on the SHD dataset.

The fact that the loss function needs to be appropriate for the learning task and that over-fitting a loss that is poorly matched to the task can lead to degradation of accuracy is well-known. Our first contribution here was to identify why trying to accommodate the original constraints of Eventprop by using the ${\cal L}_{\text{x-entropy}}$ loss was a bad idea and to extend Eventprop to allow using ${\cal L}_{\text{sum}}$ that has been widely used with BPTT and surrogate gradients. We then identified, based on our insights into the inner workings of Eventprop, that the learning-algorithm-aware loss function ${\cal L}_{\text{sum\_exp}}$ was even better.

In parallel to our work, several competing works on training SNNs for SHD and SSC datasets have been published, which use additional mechanisms and tools and achieve accuracy of up to 95.1\% on the SHD test set~\cite{dagostino2024denram,dampfhoffer2022investigating,yao2021temporal,yu2022stsc,sun2023adaptiveaxonaldelays,bittar2022surrogate,hammouamri2023learning,schöne2024scalable}.
We, therefore, investigated several augmentations of our RSNN networks and achieved the $93.5\pm 0.7$\% test accuracy reported above. This is less than some of the works listed, but some studies~\cite{schöne2024scalable,bittar2022surrogate,hammouamri2023learning} used the test set as a validation set, including for early stopping and the methodology employed in others~\cite{sun2023adaptiveaxonaldelays,yu2022stsc,yu2022stsc,yao2021temporal} is unclear whereas we here followed a strict procedure of finding parameters through 10-fold leave-one-speaker-out cross-validation and then training on the entire training set with early stopping on the training accuracy. Only then did we evaluate the performance on the test set. With a looser approach, we could have reported the highest observed test accuracy during our ablation study which was 95.5\% -- on par with the best competitors -- but that would constitute overfitting of the test set \cite{nowotny2014two}. Interestingly, our results are very similar to \cite{baronig2024advancing,higuchi2024balanced} who also used a proper validation strategy.

On the SSC dataset, we achieved an accuracy of $74.1\pm 0.9$\% whereas competitors using SNNs augmented with various additional mechanisms (convolutions~\cite{sadovsky2023speech}, adaptive neurons~\cite{yin2021accurate,bittar2022surrogate}, spiking GRU~\cite{dampfhoffer2022investigating}, delay learning~\cite{hammouamri2023learning}) have achieved test accuracies of up to  $80.29$\%. 
 It is possible that, for the SSC dataset, the additional mechanisms proposed by the competitors make a measurable difference but we speculate that surrogate gradient descent with BPPT on large timesteps (up to $25$ ms), that the competitors also used, is reducing the long-time dependencies in the data and thus simplifying the temporal credit-assignment problem. It is debatable, however, whether this still constitutes an SNN and how it would fare on neuromorphic hardware as, inevitably, a very large proportion of neurons will spike during each large timestep. It will also be interesting to see how different approaches compare on future datasets that may depend more strongly on exact spike timings so that the distinct advantage of Eventprop to allow a very large number of timesteps (and hence temporal precision), especially in the efficient GeNN implementation, may come to full fruition.

State Space Models~(SSMs) are another very recent approach to solving SSC. In parallel to the work presented here, \cite{schöne2024scalable} formulated an event-based linear SSM and achieved $88.4$\% accuracy on SSC, beating the results from SNNs, while \cite{soydan2024s7} report $88.2$\% during the revision of this manuscript with the frame-based ``S7'' deep SSM. These results suggest that the ability of SSMs to learn long-term dependencies is particularly beneficial in the SSC benchmark. However, due to their use of BPTT through JAX, these models still have very high memory requirements and need to use temporal pooling of events or large time frames. Also, while they do operate on event-based input, the discussed SSMs do not employ the same sparse, event-based computing paradigm as SNNs and, as such, it is an open question as to how they translate to neuromorphic hardware.

The Eventprop implementation in GeNN introduced in this paper is efficient and utilizes GeNN's advanced algorithms for event propagation in both the forward and backward passes. How these algorithms' time and space requirements scale with the number of events rather than timesteps can clearly be seen in our comparison with the purely clock-driven BPTT training using Spyx.
Nevertheless, a full $10$-fold cross-validation with $100$ epochs per fold still takes several hours on an A100 GPU highlighting that this type of research remains compute-intensive. 
The compute efficiency also depends on the networks operating in an appropriately sparse spiking regime. Our networks were all regularised to 14 spikes per neuron on average per trial. This is roughly in line with the 10 Hz of cortical neurons in the brain~\cite{Buzsaki2014}. After training, we observe that this target is typically achieved. Zenke and Vogels~\cite{Zenke2021} investigated this question for BPTT with surrogate gradients and found that there is a cut-off of between 10-100 spikes per trial (depending on the benchmark task) where networks fail. This is somewhat more aggressive than our choice but we have not investigated making the activity more sparse in our networks.


In conclusion, while the ability to calculate exact gradients efficiently using the EventProp method is attractive, the exact gradient is agnostic to spike creation and deletion and this can lead to learning failures in some combinations of task and loss function. In this paper, we have overcome these problems by extending EventProp to more general loss functions. We have demonstrated on the SHD benchmark how `loss shaping', i.e. choosing a bespoke loss function that induces beneficial gradient flows and learning dynamics, allows us to scale up Eventprop learning beyond proof of concept examples.
Whether loss shaping can be done in a more principled way and how it carries over to deeper networks are open questions we would like to address in the near future. 
We are also planning to combine Eventprop with advanced network architectures to overcome the known limitations of simple recurrent hidden layers in learning long-time dependencies. Further directions include an Eventprop algorithm for learning delays and more complex neuron models such as resonate-and-fire neurons that may link back to the successful state space models discussed above.

\section{Materials and Methods}
\subsection{Phantom spike regularisation}
For the time to first spike loss function ${\cal L}_{\text{time}}$, there is a risk that if the correct output neuron does not spike despite the loss term that tries to push its spikes forward in time, there will be no valid gradient to follow. To address this issue we introduced `phantom spikes' so that if the correct neuron does not spike during a trial, a regularisation loss term is added to the $\lambda_V$ dynamical equation as if the neuron had spiked at time $T$.

\subsection{Regularisation in the hidden layer}

When hidden layer neurons spike too frequently or cease to spike, network performance degrades. We therefore would like to introduce a regularisation term to the loss function, as in \cite{Zenke2021} that penalises derivations from a target firing rate,
\begin{eqnarray}
  {\cal L}_{\text{reg}}= \frac{1}{2} k_{\text{reg}} \sum_{l=1}^{N_{\text{hidden}}} \left(\left(\frac{1}{N_{\text{batch}}} \sum_{n=1}^{N_{\text{batch}}} n^{\text{spike},n}_l\right) - \nu_{\text{hidden}}\right)^2
\end{eqnarray}
where $n^{\text{spike},n}_l$ denotes the number of spikes in hidden neuron $l$ in trial $n$ and $\nu_{\text{hidden}}$ represents the target number of spikes in a trial. For example, in the SHD experiments, $\nu_{\text{hidden}}=14$ corresponds to a 14 Hz target firing rate in a 1000 ms trial. $k_{\text{reg}}$ is a free parameter scaling the strength of the regularisation term. As the Eventprop formalism cannot be applied to this loss term because it cannot be expressed in a meaningful way as a function of spike times or of the membrane potential, we instead use the heuristic jumps of 
\begin{eqnarray}
\lambda_{V,l}^- = \lambda_{V,l}^+ - \frac{k_{\text{reg}}}{N_{\text{batch}}} \left(\left(\frac{1}{N_{\text{batch}}} \sum_{n=1}^{N_{\text{batch}}} n^{\text{spike},n}_l\right) - \nu_{\text{hidden}}\right)
\end{eqnarray}
at recorded spikes of hidden neuron $l$ during the backward pass. We also trialled other regularisation terms, for instance involving per-trial spike counts but found this simple heuristic to be the most useful.

\subsection{Dropout and Noise}
We experimented with dropout in the input and hidden layer where input spikes occurred only with a probability $1-p^\text{drop}_\text{in} < 1$ for each input spike and spikes in the hidden layer occurred only with probability $1 - p^\text{drop}_\text{hid}$ upon a threshold crossing. We did not observe improvements in the classification accuracy of SHD and hence did not include these mechanisms in the later parts of this work. We also experimented with additive membrane potential noise in the hidden layer, which also did not have positive effects and was subsequently omitted from the analysis.

The parameters used for the first assessments of MNIST (Figure S1) and SHD with different loss functions (\ref{fig:SHD}) are detailed in tables \ref{table2}, \ref{table3}, \ref{table4}.

\begin{table}
  \caption{Model parameters for MNIST and loss function comparison on SHD (Fig \ref{fig:SHD})}
  \label{table2}
  \begin{tabular}{lp{6cm}ll}
    \toprule
    Name & Description & Value MNIST & Value base SHD \\
    \midrule
    $\tau_{\text{mem}}$ (ms) & timescale of membrane potential & $20$ & table \ref{table3} \\
    $\tau_{\text{syn}}$ (ms) & synaptic timescale & $5$ & table \ref{table3} \\
    $T$ (ms) &  trial duration & $20$ & $1400$ \\
    $\mu_{i\_h}$ & Mean initial weight value input to hidden & $0.045$ & $0.03$ \\
    $\sigma_{i\_h}$ & Standard deviation of initial weight value input to hidden & $0.045$ & $0.01$ \\ 
    $N_{\text{hidden}}$ & Number of hidden neurons & $128$ & $256$ \\
    $\mu_{h\_o}$ & Mean initial weight value hidden to output & $\!\!\!\!\left\{\begin{array}{ll} \!\!\! 0.9 & \!\! \text{if } {\cal L}_{\text{time}} \\ \!\!\! 0.2  & \!\! \text{o.w.} \end{array}\right.$ & table \ref{table3} \\
    $\sigma_{h\_o}$ & Standard deviation of initial value hidden to output & $\!\!\!\!\left\{\begin{array}{ll} \!\!\! 0.03 & \!\! \text{if } {\cal L}_{\text{time}} \\ \!\!\! 0.37 & \!\! \text{o.w.} \end{array}\right.$ & table \ref{table3} \\
    $\mu_{h\_h}$ & standard deviation of initial weight value hidden to hidden & -- & $0$ \\
    $\sigma_{h\_h}$ & standard deviation of initial weight value hidden to hidden & -- & $0.02$ \\
    $p_{\text{drop}}$ & Dropout probability for input spikes & $0.2$ & $0$ \\
    $\nu_{\text{hidden}}$ & target hidden spike number for regularisation & $4$ if ${\cal L}_{\text{time}}$ & $14$ \\
    $\eta$ & Learning rate & $\!\!\!\!\left\{\begin{array}{ll} \!\!\! 5\cdot 10^{-3} & \!\! \text{if } {\cal L}_{\text{time}} \\ \!\!\! 10^{-2} & \!\! \text{o.w.} \end{array}\right.$ & table \ref{table3} \\
    $\tau_0$ (ms) & Parameter of timing loss ${\cal L}_{\text{time}}$ & $1$ & $1$ \\
    $\tau_1$ (ms) & Parameter of timing loss ${\cal L}_{\text{time}}$ & $3$ & $100$ \\
    $\alpha$ & Parameter of timing loss ${\cal L}_{\text{time}}$ & $3.6\cdot 10^{-4}$ & $5 \cdot 10^{-5}$ \\
    $N_{\text{epoch}}$ & number of training epochs & $50$ & $300$ \\
    \bottomrule
  \end{tabular}
\end{table}

\begin{table}
  \caption{Specific parameters of different base SHD models}
  \label{table3}
  \begin{tabular}{lllllllll}
    \toprule
    Parameter & \multicolumn{8}{c}{Models} \\
    \midrule    
    loss & \multicolumn{2}{c}{${\cal L}_{\text{sum}}$} & \multicolumn{2}{c}{${\cal L}_{\text{sum\_exp}}$}  &  \multicolumn{2}{c}{${\cal L}_{\text{time}}$} & \multicolumn{2}{c}{${\cal L}_{\text{max}}$} \\
    architecture & ffwd & recur & ffwd & recur & ffwd & recur & ffwd & recur \\
    $\tau_{\text{mem}}$ & 20 & 20 & 40 & 40 & 40 & 40 & 40 & 40 \\
    $\tau_{\text{syn}}$ & 10 & 10 & 5 & 5 & 5 & 5 & 10 & 10 \\
    $k_{\text{reg}}$ & $10^{-12}$ & $10^{-12}$ & $10^{-10}$ & $10^{-9}$ & $10^{-07}$ & $10^{-07}$ & $5 \cdot 10^{-9}$ & $5 \cdot 10^{-9}$ \\
    $\mu_{h\_o}$ & $0$ & $0$ & $0$ & $0$ & $1.2$ & $1.2$ & $0$ & $0$ \\
    $\sigma_{h\_o}$ & $0.03$ & $0.03$  & $0.03$ & $0.03$ & $0.6$ & $0.6$ & $0.03$ & $0.03$ \\
    $\eta$ & $0.002$ & $0.002$ & $0.001$ & $0.001$ & $0.001$ & $0.001$ & $0.002$ & $0.002$ \\
    \bottomrule 
  \end{tabular}
\end{table}

\begin{table}
  \caption{Model parameters common to all experiments}
  \label{table4}
  \begin{tabular}{lll}
    \toprule
    Name & Description & Value \\
    \midrule
    $\vartheta$  & firing threshold & $1$ \\
    $V_{\text{reset}}$ & reset for membrane potential & $0$ \\
    $\beta_1$ & Adam optimiser parameter & $0.9$ \\
    $\beta_2$ & Adam optimiser parameter & $0.999$ \\
    $\epsilon$ & Adam parameter & $10^{-8}$ \\
    $N_{\text{batch}}$ & mini-batch size & $32$ \\
    $\Delta t$ (ms)& simulation timestep & $1$ \\
    \bottomrule
  \end{tabular}
\end{table}

\begin{table}
 \caption{Parameters in final runs (Figures \ref{fig:final_SHD},\ref{fig:final_SSC}), if different from tables \ref{table2}, \ref{table3}}
 \label{table5}
\begin{tabular}{llll}
    \toprule
     Name & Description & Value \\
    \midrule
    $\tau_{\text{mem}}$ (ms) & membrane time constant & $20$ \\
    $\tau_{\text{syn}}$ (ms) & synapse time constant & $5$ \\
    $T$ (ms) & trial duration & $1000$ \\
    $k_{\text{reg}}$ & regularisation strength & individual \\
    $\mu_{h\_o}$ & mean of initial hidden-output weights & $0$ \\
    $\sigma_{h\_o}$ & standard deviation of initial hidden-output weights & $0.03$ \\
    $\eta$ & initial learning rate & $0.001$ \\
    $f_{\text{shift}}$ & amplitude of "frequency shift" augmentation & $40$ \\
    $N_{\text{delay}}$ & Number of delayed copies of input neurons & $10$ \\
    $t_{\text{delay}}$ (ms) & Delay between the separate input copies & $30$ \\
    $p_{\text{blend}}$ & Probability to accept a spike candidate during blending & $0.5$ \\
    \bottomrule
\end{tabular}
\end{table}

\subsection{Augmentations}
We initially investigated four types of input augmentation to lessen the detrimental effects of over-fitting.
\begin{enumerate}
    \item The ID jitter augmentation was implemented as in \cite{Cramer2022}. For each input spike, we added a ${\cal N}(0,\sigma_u)$ distributed random number to the index $i$ of the active neuron, rounded to an integer and created the spike in the corresponding neuron instead.
    \item In the random dilation augmentation, we rescaled the spike times of each input pattern homogeneously by a factor random factor $k^{\text{scale}}$ drawn uniformly from $[k^{\text{scale}}_{\text{min}}, k^{\text{scale}}_{\text{max}}]$.
    \item In the random shift augmentation we globally shifted the input spikes of each digit across input neurons by a distance $k^{\text{shift}}$ uniformly drawn from $[-f_{\text{shift}}, f_{\text{shift}}]$, rounded to the nearest integer.
    \item In the blend augmentation, the spikes from two randomly chosen input patterns $x_1$ and $x_2$ of the same class are ``blended'' into a new input pattern by including spikes from $x_1$ with probability $p_1$ and spikes from $x_2$ with probability $p_2$. We initially trialled different combinations of $p_1 + p_2 = 1$, to restrict blending to examples from the same speaker, and blending three inputs but eventually settled on blending 2 inputs, potentially from different speakers, with $p_1= p_2 = 0.5$. Before blending the inputs are aligned to their centre of mass along the time axis. We generated the same number of additional inputs as there were in the data set originally.
\end{enumerate}
However, we observed that only the random shift augmentation and the blend augmentation improved generalisation noticeably and we conducted the remainder of the research only with these two augmentation types.

\subsection{Silent neurons}
In the final version of the model, we implemented a safeguard against hidden neurons becoming completely silent. Whenever a hidden neuron does not fire a single spike during an entire epoch, we add $\Delta g = 0.002$ to all of its incoming synapses. This repeats each epoch the neuron is silent but, as soon as at least one spike is fired, normal synaptic updates implementing stochastic gradient descent with Eventprop resume.

\subsection{Learning rate ease-in}
It is very difficult to initialise synapses so that their activity is well-suited for learning. In particular, it is often the case that the initial synaptic weights cause inappropriate levels of activity which in turn causes very large synaptic updates from the regularisation loss terms. This can lead to learning failure. The phenomenology of this problem is typically an immediate shutdown of all activity in the first few mini-batches. To avoid the problem we ``ease in'' the learning rate, starting with $\eta_0 = 10^{-3}\cdot \eta$ and then increasing the learning rate by a factor $1.05$ each mini-batch until the full desired rate $\eta$ is reached. 

\subsection{Learning rate schedule}
In the final version of the network, we also use a learning rate schedule driven by two exponentially weighted moving averages of either the validation accuracy (in cross-validation on SHD or train/validation runs on SSC) or of the training accuracy (when running train/test on SHD),
\begin{eqnarray}
    m_{\text{fast}} (n+1) &=& 0.8 \, m_{\text{fast}}(n)+0.2 \, x(n) \\
    m_{\text{slow}} (n+1) &=& 0.85 \, m_{\text{slow}}(n)+0.15 \, x(n) 
\end{eqnarray}
where $n$ is the epoch index and $x(n)$ is the average accuracy in the epoch. Whenever $m_{\text{fast}} < m_{\text{slow}}$, i.e. the accuracy decreased again, and at least $50$ epochs have passed since the last learning rate change, the learning rate is reduced to half its value.

\subsection{Heterogeneous timescale initialisation}
When using heterogeneous values for $\tau_{\text{mem}}$ and $\tau_{\text{syn}}$, we initialised the time scales from third-order gamma distributions following \cite{perez2021neural},
\begin{eqnarray}
   \tau_{\text{mem}} \sim \Gamma(3,\bar{\tau}_{\text{mem}}/3) \\
   \tau_{\text{syn}} \sim \Gamma(3,\bar{\tau}_{\text{syn}}/3)
\end{eqnarray}
where $\bar{\tau}_x$ are the corresponding homogeneous values used. We clipped the values to  $\tau_{\text{mem}} \in [3 \text{ms}, 3\bar{\tau}_{\text{mem}}]$ and $\tau_{\text{syn}} \in [1 \text{ms}, 3\bar{\tau}_{\text{syn}}]$.

\subsection{Learning timescales}
When employing learning of timescales $\tau_{\text{mem}}$ and $\tau_{\text{syn}}$, we use the gradient equations (\ref{eqn:gradtaum}) and (\ref{eqn:gradtausyn}) derived in Appendix \ref{sec:taulearn} and clip values to $\tau_{\text{mem}} \geq 3$ms and $\tau_{\text{syn}} \geq 1$ms. There was no upper limit for the timescales during learning.

\subsection{Validation methodology in ablation experiments}
\emph{SHD:} For each possible combination of parameters, we determined the optimal regularisation strength $k_{\text{reg}}$ from five candidate values based on leave-one-speaker-out cross-validation with early stopping on the training error. We left all other uninvestigated parameters the same as in our reference solution and used the identified best regularisation strength for eight independent test runs with different random seeds. For these test runs, we trained on the full SHD training set and used early stopping on the epoch with the lowest training error. We then report the test accuracy for the network with the weights from this epoch.

\emph{SSC:}
We trained the networks for each parameter combination, five candidate regularisation strengths $k_{\text{reg}}$ and two different random seeds on the training set and performed early stopping on the validation set. We then chose the regularisation strength for each parameter combination that led to the best average validation accuracy across the two seeds and used the corresponding weights (and $\tau$'s if tau-learning was enabled) for inference on the test set.
For each of the best regularisation choices we then ran an additional $6$ repeats of training and testing to be able to report 8 independent measurements of the test error.

The parameters used for the in-depth assessment of SHD and SSC (Figures \ref{fig:final_SHD},\ref{fig:final_SSC}) were as for the base SHD model with ${\cal L}_{\text{sum\_exp}}$, except that we cropped the trials to $T = 1000$ ms, and used $\tau_{\text{mem}} = 20$ ms, $N_{\text{epoch}} =100$ (no "blend") or $50$ ("blend"), and the candidate values $k_{\text{reg}} \in \{5 \cdot 10^{-11}, 2.5 \cdot 10^{-10}, 5\cdot 10^-{10}, 2.5\cdot 10^{-9}, 5 \cdot 10^{-9}\}$ in the leave-one-speaker-out cross-validation.

\subsection{Benchmarking against back-propagation through time}
For our benchmarking we used Spyx 0.1.19 and JAX 0.4.26 and modified the SHD classifier model developed by Heckel et al.~\cite{heckel_spyx_2024} to use recurrently connected LIF neurons with exponential synapses (``RCuBaLIF'') to match our EventProp models.
In order to focus on the most relevant core training, we compared the Spyx model to a non-augmented EventProp model with homogeneous timescales and no silent neuron up-weighting.
When measuring peak memory utilisation, we performed separate runs with the ``XLA\_PYTHON\_CLIENT\_ALLOCATOR`` environment variable set to ``platform`` and measured memory usage using the ``nvidia-smi`` command line tool.

In both the GeNN and Spyx simulations, the entire SHD dataset is uploaded to GPU memory at the start of training. However, they use different data structures. To fairly compare the simulation memory usage, we therefore subtract the size of these data structures from the total GPU memory usage for both simulators.
When measuring training time we excluded the time taken to load the SHD dataset but made sure to include the time taken to generate code and compile it (GeNN) or Just-In-Time compile the JAX model (Spyx).

\subsection{Implementation details}
We have used the GeNN simulator version 4.9.0 \cite{Yavuz2016,genn_github} through the PyGeNN interface \cite{Knight2021} for this research. Spike times and the derivative of $V$ at the spike times are buffered in memory during the forward pass, with buffers that can hold up to $1500$ spikes per two trials. The EventProp backward pass is implemented with additional neuron variables $\lambda_V$ and $\lambda_I$ within the neuron {\tt sim\_code} code snippet. For efficiency, we ran the forward pass of mini-batch $i$ and the backward pass of mini-batch $i-1$ simultaneously in the same neurons. During mini-batch $0$ no backward pass is run and the backward pass of the last mini-batch in each epoch is not simulated. In initial experiments, we observed no measurable difference other than reduced runtime when compared to properly interleaved forward-only and backward-only simulations.

We obtained the SHD and SSC datasets from the tonic library \cite{lenz_gregor_2021_5079802}.
During training the inputs were presented in a random order during each epoch. For the weight updates, we used the Adam optimizer \cite{Kingma2015}. The parameters of the simulations are detailed in tables \ref{table2} and \ref{table3}. The simulation code is published on Github \cite{genn_eventprop}.
The data underlying Figures \ref{fig:SHD}, \ref{fig:final_SHD}, \ref{fig:final_SSC}, S2 and  S3 are available on figshare (https://figshare.com/s/ec4841f808ff707bed66). Figures \ref{fig:axe1} and S1 can be recreated with minimal effort from scripts in the Github repository.

Simulations were run on local workstations with NVIDIA GeForce RTX 3080 and A5000 GPUs, on the JADE 2 GPU cluster equipped with NVIDIA V100 GPUs and the JUWELS-Booster system at the J\"ulich Supercomputer Centre, equipped with NVIDIA A100 GPUs. All runtimes are reported for A100 GPUs.\

\section{References}


\providecommand{\newblock}{}

\section{Acknowledgments}
We thank Christian Pehle for his feedback on Appendix \ref{sec:taulearn} and Christopher L Buckley for helpful comments on the manuscript. \\

\noindent
{\bf Funding:} This work was funded by the EPSRC (Brains on Board project, Grant Number EP/P006094/1, ActiveAI project, Grant Number EP/S030964/1, Unlocking spiking neural networks for machine learning research, Grant Number EP/V052241/1) and the European Union's Horizon 2020 research and innovation programme under Grant Agreement 945539 (HBP SGA3). Additionally, we gratefully acknowledge the Gauss Centre for Supercomputing e.V. (www.gauss-centre.eu) for funding this project by providing computing time through the John von Neumann Institute for Computing (NIC) on the GCS Supercomputer JUWELS at Jülich Supercomputing Centre (JSC); and the JADE2 consortium funded by the EPSRC (EP/T022205/1) for compute time on their systems. \\

\noindent
{\bf Author contributions:} \\
Conceptualization: TN \\
Methodology: TN, JCK \\
Investigation: TN, JCK \\
Visualization: TN, JCK \\
Code---original: TN \\
Code---review \& refactoring: TN, JT \\
Writing---original draft: TN \\
Writing---review \& editing: TN, JCK, JT \\

\noindent
{\bf Competing interests:} The authors declare that there are no competing interests. \\

\noindent
{\bf Data and materials availability:} The code underlying the results presented is available at \url{https://github.com/tnowotny/genn_eventprop}\\
The data underlying the figures is available on the University of Sussex figshare repository at \url{https://figshare.com/s/ec4841f808ff707bed66}. This work was performed with GeNN 4.9.0 \url{https://zenodo.org/records/8430715}.

\clearpage

\appendix
\section{Full derivation of extended Eventprop}
\label{sec:extended_Eventprop}
In this Appendix we present the detailed derivation for extending the EventProp algorithm to losses of the shape
\begin{eqnarray}
    {\cal L}_F = F\left(\textstyle \int_0^T l_V(V(t),t) \, dt\right), \label{eqn:loss_gen_2}
\end{eqnarray}
where $F$ is a differentiable function and $l_V$ can be vector-valued, e.g. $l_V = V$ as in the loss functions used in the main body of the paper. 

Using the chain rule, we can calculate
\begin{eqnarray}
    \frac{d{\cal L}_F}{d w_{ji}} = \frac{\partial F}{\partial \left(\int_0^T l_V \, dt\right)} \cdot \frac{d \left(\int_0^T l_V \, dt\right)}{d w_{ji}} = \sum_{n} \frac{\partial F}{\partial \left(\int_0^T l_V^{n} \, dt\right)}  \frac{d \left(\int_0^T l_V^{n} \, dt\right)}{d w_{ji}}
\end{eqnarray}
where $n$ labels the components of the vector-valued $l_V$, for instance $l_V$= $V_{\text{output}}$ and $n$ labels the output neurons (see below).
The integrals $\int_0^T l_V^n(t) dt$ are of the shape of a classic Eventprop loss function, so we can use the Eventprop algorithm to calculate
\begin{eqnarray}
    \frac{d \left(\int_0^T l_V^n \, dt\right)}{d w_{ji}} = -\tau_{\text{syn}} \sum_{t \in t_{\text{spike}}(i)} \lambda_{I,j}^n(t),
\end{eqnarray}
where
\begin{eqnarray}
    \tau_{\text{syn}} {\lambda_{I,j}^n}' = -\lambda_{I,j}^n + \lambda_{V,j}^n \\
    \tau_{\text{mem}} {\lambda_{V,j}^n}' = -\lambda_{V,j}^n - \frac{\partial l_V^n}{\partial V_j}
\end{eqnarray}
With this in mind, we can then calculate the gradient of the loss function (\ref{eqn:loss_gen_2}):
\begin{eqnarray}
    \frac{d {\cal L}_F}{d w_{ji}} &=& -\tau_{\text{syn}} \sum_{n} \frac{\partial F}{\partial (\int_0^T l_V^n dt)} \sum_{t \in t_{\text{spike}}(i)} \lambda_{I,j}^{n}(t) \\
      &=& -\tau_{\text{syn}} \sum_{t \in t_{\text{spike}}(i)} \sum_{n} \frac{\partial F}{\partial (\int l_V^n dt)}  \lambda_{I,j}^{n}(t) 
        = -\tau_{\text{syn}} \sum_{t \in t_{\text{spike}}(i)} \tilde{\lambda}_{I,j} (t) \label{eqn:deriv}
\end{eqnarray}
where we have defined
\begin{eqnarray}
    \tilde{\lambda}_{I,j} (t):= \sum_{n} \frac{\partial F}{\partial (\int l_V^n dt)}  \lambda_{I,j}^{n}(t).
\end{eqnarray}
We can then derive dynamics for $\tilde{\lambda}_{I,j}$ by simply using this definition and noting that $\frac{\partial F}{\partial (\int l_V^n dt)}$ does not depend on $t$,
\begin{eqnarray}
    \tau_{\text{syn}} \tilde{\lambda}_{I,j}' &=& \sum_{n} \frac{\partial F}{\partial (\int l_V^n dt)}  \tau_{\text{syn}} {\lambda_{I,j}^{n}}' \\
    &=& \sum_{n} \frac{\partial F}{\partial (\int l_V^n dt)} \left(-  \lambda_{I,j}^{n} +  \lambda_{V,j}^{n}\right) 
    = -\tilde{\lambda}_{I,j} + \tilde{\lambda}_{V,j}, \label{eqn:lambda_I_dyn}
\end{eqnarray}
where we defined
\begin{eqnarray}
    \tilde{\lambda}_{V,j}:= \sum_{n} \frac{\partial F}{\partial (\int l_V^n dt)} \lambda_{V,j}^{n}.
\end{eqnarray}
That implies the dynamics
\begin{eqnarray}
    \tau_{\text{mem}} \tilde{\lambda}_{V,j}' &=&  \sum_{n} \frac{\partial F}{\partial (\int l_V^n dt)} \tau_{\text{mem}} {\lambda_{V,j}^{n}}' 
    = - \tilde{\lambda}_{V,j} - \sum_{n} \frac{\partial F}{\partial (\int l_V^n dt)} \frac{\partial l_V^n}{\partial V_j}.\label{eqn:lambda_V_dyn}
\end{eqnarray}
In this fashion, we have recovered an Eventprop algorithm to calculate the gradient of the general loss function (\ref{eqn:loss_gen_2}) by using equations (\ref{eqn:deriv}),(\ref{eqn:lambda_I_dyn}), and (\ref{eqn:lambda_V_dyn}). Remarkably, this algorithm is exactly as the original Eventprop algorithm except for the slightly more complex driving term for $\tilde{\lambda}'_{V,j}$ in equation (\ref{eqn:lambda_V_dyn}).

\section{Extended Eventprop for specific loss functions}
\label{sec:specific_loss}
For the concrete example of the loss function (\ref{eqn:sumloss}), we have $l_V= (V_n^m)$, i.e. the vector of output voltages of output neuron $n$ in each batch $m$. With respect to the mini-batch summation we typically calculate the gradient as the sum of the individual gradients for each trial in the mini-batch,
\begin{eqnarray}
    \frac{d {\cal L}}{dw_{ji}} &=& \frac{1}{N_{\text{batch}}} \sum_{m=1}^{N_{\text{batch}}} \frac{d}{d w_{ji}} \left(-\log \frac{\exp\left(\int_0^T V_{l(m)}^m(t) dt\right)}{\sum_{k=1}^{N_{\text{out}}} \exp\left(\int_0^T V_{k}^m(t) dt\right)}\right) \\
    &=& \frac{1}{N_{\text{batch}}} \sum_{m=1}^{N_{\text{batch}}} \frac{dF^m}{d w_{ji}}, \label{lossadd}
\end{eqnarray}
with 
\begin{eqnarray}
    F^m({\textstyle\int_0^T V dt})= -\log \frac{\exp\left(\int_0^T V_{l(m)}^m(t) dt\right)}{\sum_{k=1}^{N_{\text{out}}} \exp\left(\int_0^T V_{k}^m(t) dt\right)}
\end{eqnarray} 
and for each trial $m$ and output neuron $n$ we get
\begin{eqnarray}
    \frac{\partial F^m}{\partial (\int_0^T V_n^m dt)} =  
      - \delta_{n,l(m)} + \frac{\exp(\int V_n^m dt)}{\sum_{k=1}^{N_{\text{class}}} \exp(\int_0^T V_k^m dt)} 
\end{eqnarray}
and $\frac{\partial l_V}{\partial V_j^m} = \frac{\partial V_n^m}{\partial V_j^m} = \delta_{jn}$. We, therefore, can formulate the Eventprop scheme
\begin{eqnarray}
    \frac{d F^m}{d w_{ji}} &=& -\tau_{\text{syn}} \sum_{\{t_{\text{spike}}(i)\}} \tilde{\lambda}_{I,j}^m(t_{\text{spike}}) \\
    \tau_{\text{syn}} \tilde{\lambda}_{I,j}^m{}' &=& -\tilde{\lambda}_{I,j}^m + \tilde{\lambda}_{V,j}^m \\
    \tau_{\text{mem}} \tilde{\lambda}_{V,j}^m{}' &=& - \tilde{\lambda}_{V,j}^m + \delta_{j,l(m)} -\frac{\exp(\int V_j^m dt)}{\sum_{k=1}^{N_{\text{class}}} \exp(\int_0^T V_k^m dt)} . \nonumber
\end{eqnarray}
i.e. there is a positive contribution $1$ for each correct output neuron and the negative fraction of exponentials for all output neurons. All other neurons do not enter the loss function directly and the loss propagates as normal from the output neurons through the $W^T(\lambda_V^+-\lambda_I)$ terms. The final loss is then added up according to (\ref{lossadd}).

Another typical loss function for classification works with the maxima of the voltages of the non-spiking output neurons
\begin{eqnarray}
    {\mathcal L_{\text{max}}} = - \frac{1}{N_{\text{batch}}} \sum_{m=1}^{N_{\text{batch}}} \log \frac{\exp\left( \max_{[0,T]} V_{l(m)}^m(t) \right)}{ \sum_{k=1}^{N_{\text{class}}} \exp\left(\max_{[0,T]} V_{k}^m(t) \right)}
\end{eqnarray}
As already argued in \cite{Wunderlich2021}, we can rewrite 
\begin{eqnarray}
\max_{[0,T]} V_{n}^m(t) = \int_0^T V_n^m (t) \delta(t - t_{\text{max},n}^m) \, dt, 
\end{eqnarray}
where $\delta$ is the Dirac distribution and $t_{\text{max}}$ the time when the maximum $V$ was observed. Then we can use arguments as above to find the Eventprop scheme
\begin{eqnarray}
    \frac{d{F^m}}{d w_{ji}} &=& -\tau_{\text{syn}} \sum_{ t \in t_{\text{spike}}(i)} \tilde{\lambda}_{I,j}^m(t) \\
    \tau_{\text{syn}} \tilde{\lambda}_{I,j}^m{}' &=& -\tilde{\lambda}_{I,j}^m + \tilde{\lambda}_{V,j}^m \\
    \tau_{\text{mem}} \tilde{\lambda}_{V,j}^m{}' &=& - \tilde{\lambda}_{V,j}^m + \left(\delta_{j,l(m)} -\frac{\exp\left(\max_{[0,T]} V_j^m\right)}{\sum_{k=1}^{N_{\text{class}}} \exp\left( \max_{[0,T]} V_k^m \right)} \right) \delta\left(t-t_{\text{max},j}^m\right), \nonumber
\end{eqnarray}
i.e. we have jumps at the times where the maximum voltages in each of the neurons occurred, with a magnitude determined by the combination of the maximum voltages of all output neurons.

\newcommand{\liexp}{\exp\left(-t_{i,l(i)}/\tau_0\right)}
\newcommand{\liexpb}{\exp\left(t_{i,l(i)}/\tau_1\right)}
\newcommand{\smexp}{\sum_{k=1}^{N_{\text{class}}} \exp\left(-t_{i,k}/\tau_0\right)}
Finally, the spike-time based loss was defined by Wunderlich and Pehle \cite{Wunderlich2021} as,
\begin{eqnarray}
      {\cal L}= & &- \frac{1}{N_{\text{batch}}}
  \Bigg[\sum_{i=1}^{N_{\text{batch}}} \log \left[
      \frac{\liexp}{\smexp}\right] \nonumber \\
      & &- \alpha
    \left[\liexpb -1 \right] \Bigg],
\end{eqnarray}
which is a loss function of the form  $\sum_k l_p(t_k)$. The Eventprop algorithm is hence driven by the jumps 
\begin{eqnarray}
    \frac{1}{\tau_{\text{mem}}(\dot{V}^-)_{n(k)}} \frac{\partial l_p}{\partial t_k}. 
\end{eqnarray}
For a spike of the correct output neuron $l(i)$ in trial $i$, we have
\begin{eqnarray}
    \frac{\partial l_p}{\partial t_{i,l(i)}} = & & \frac{1}{N_{\text{batch}}} \Bigg[ \frac{1}{\tau_0} \left(1 - \frac{\liexp}{\smexp}\right) \nonumber \\
    & & + \frac{\alpha}{\tau_1} \liexpb \Bigg]
\end{eqnarray}
and for the other output neurons $j \neq l(i)$,
\begin{eqnarray}
    \frac{\partial l_p}{\partial t_{i,j}} = \frac{1}{N_{\text{batch}}} \left[ -\frac{1}{\tau_0} \frac{\exp\left(-t_{i,j} / \tau_0\right)}{\smexp}\right]
\end{eqnarray}

\section{Eventprop for neuron properties}
\label{sec:taulearn}
As published, Eventprop is formulated to calculate the exact gradient with respect to synaptic weights. However, there are indications that in can be beneficial if neuron properties are heterogeneous and even trainable \cite{perez2021neural}. In this section of the Appendix we show how the Eventprop formalism can be extended to include gradient calculations for the neuron membrane time constant.

We derive the scheme by repeating the work presented in \cite{Wunderlich2021} for the derivative $\frac{d {\cal L}}{d \tau_{\text{mem},i}}$. We start with the same standard Eventprop loss function, equation (1) in \cite{Wunderlich2021} and similar dynamic equations (25), except that $\tau_{\text{mem}}$ is now individual to each neuron, i.e. $\tau_{\text{mem}} \in {\mathbb R}^n$,
\begin{eqnarray}
    f_V \equiv \tau_{\text{mem}} \odot \dot{V} + V - I = 0
\end{eqnarray}
where $\odot$ denotes the Hadamard product.

Then we calculate similarly to their equation (26) 
\begin{eqnarray}
    \frac{d \cal L}{d \taumi} = \frac{d}{d \taumi} \left[
        l_p (t^{\text{post}}) + \sum_{k=0}^{N_{\text{post}}}
        \int_{t_k^{\text{post}}}^{t_{k+1}^{\text{post}}} [l_V(V,t) +
          \lambda_V \cdot f_V + \lambda_I \cdot f_I ] dt \right]  \nonumber
\end{eqnarray}
but now, as we are taking the derivative with respect to $\taumi$, equation (27a) yields
\begin{eqnarray}
    \frac{\partial f_V}{\partial \taumi} = e_i \odot \dot{V} + \tau_{\text{mem}} \odot
    \frac{d}{dt} \frac{\partial V}{\partial \taumi} + \frac{\partial
      V}{\partial \taumi} - \frac{\partial I}{\partial
      \taumi}.
\end{eqnarray}
This means the remainder of the proof can go ahead analogous to the original except for the extra term $\lambda_{V,i}\dot{V}_i$ within the integral between spike times (28).

It is straightforward to follow that the analogues of the jump rules up to equation (44) all carry through analogously for the derivative with respect to $\taumi$. Then, however, when taking the derivative with respect to $\taumi$ instead of $w_{ji}$ in equation (46) the term $\delta_{in}\delta_{jm}$ does not arise (the partial derivative of $w_{nm}$ with respect to $\taumi$ is zero). So, when we put it all together and apply the equivalent smart choices for the jumps of the adjoint variables, we can eliminate all terms in the equivalent of $\xi_k$ except for the additional $\lambda_{V,i}\dot{V}_i$, so that
\begin{eqnarray}
\frac{d \cal L}{d \taumi} = \sum_{k=0}^{N_{\text{post}}} \left[
      \int_{t_k^{\text{post}}}^{t_{k+1}^{\text{post}}} \lambda_{V,i}\dot{V}_i \, dt \right] 
      = \int_0^T \lambda_{V,i}\dot{V}_i \, dt . \label{eqn:gradtaum}
\end{eqnarray}

 Similarly, if we are looking for the derivative with respect to $\tausi$ after generalising (25b) in \cite{Wunderlich2021} to
 \begin{eqnarray}
     f_I \equiv \tau_{\text{syn}} \odot \dot{I} + I = 0
 \end{eqnarray}
with $\tau_{\text{syn}} \in {\mathbb R}^n$, we arrive at
\begin{eqnarray}
    \frac{d \cal L}{d \tausi} = \int_0^T \lambda_{I,i} \dot{I}_i \, dt  \label{eqn:gradtausyn}
\end{eqnarray}
Note that in this approach $\tausi$ is interpreted as a property of the post-synaptic neuron, i.e. while different post-synaptic neurons can have different synaptic timescales, this is per neuron and all synaptic currents entering the same postsynaptic neuron will decay with the same timescale.

\begin{table}
\caption{Extended Eventprop algorithm for a wider class of loss functions and including $\tau$ learning}
\begin{tabular}{lll}
  \toprule
  \multicolumn{3}{l}{{\bf Loss function:} ${\cal L}_F = l_p(t^{\text{post}}) + F\left(\textstyle \int_0^T l_V(V(t),t) \, dt\right)$, where $l_V$ can be a vector} \\
  \midrule
  {\bf Free dynamics} & {\bf Transition } & {\bf Jumps at
    transition} \\
    & {\bf condition} & \\
  \midrule 
  Forward: \\
 $\tau_{\text{mem}} \frac{d}{dt} V = -V + I$ & $(V)_n - \vartheta = 0$,
   &
  $(V^+)_n = 0$ \\
  $\tau_{\text{syn}} \frac{d}{dt} I= -I$ & $\big(\dot{V}\big)_n \neq 0$ & $I^+= I^- + W
  e_n$ \\
  \midrule
    Backward: \\
  $\tau_{\text{mem}} \lambda_V' = - \lambda_{V} - \frac{\partial F}{\partial (\int l_V dt)} \frac{\partial l_V}{\partial V}$ & $t-t_k = 0$ &
  $(\lambda_V^-)_{n(k)} = (\lambda_V^+)_{n(k)} +
  \frac{1}{\tau_{\text{mem}} (\dot{V}^-)_{n(k)}} \Big[
      \vartheta (\lambda_V^+)_{n(k)}$ \\
     $\tau_{\text{syn}} \lambda_I' = -\lambda_I + \lambda_V$ & & $\hphantom{(\lambda_V^-)_{n(k)} = } + \left(W^T
      (\lambda_V^+ - \lambda_I)\right)_{n(k)} + \frac{\partial
        l_p}{\partial t_k} + l_V^- - l_V^+ \Big]$ \\
  \midrule
    {\bf Gradient of the loss:} \\
    \multicolumn{3}{l}{$\frac{d {\cal L}}{d w_{ji}} =-\tau_{\text{syn}} \sum_{t \in t_{\text{spike}}(i)} \lambda_{I,j} (t) \quad\quad$ 
    $\frac{d \cal L}{d \taumi} = \int_0^T \lambda_{V,i}\dot{V}_i \, dt \quad\quad$ 
    $\frac{d \cal L}{d \tausi} = \int_0^T \lambda_{I,i} \dot{I}_i \, dt$} \\
  \bottomrule
\end{tabular} 
\end{table}

\section*{Supplementary material}
\setcounter{figure}{0}
\makeatletter 
\renewcommand{\thefigure}{S\@arabic\c@figure}
\makeatother
\begin{figure}[h]
    \centering
    \includegraphics[width=\textwidth]{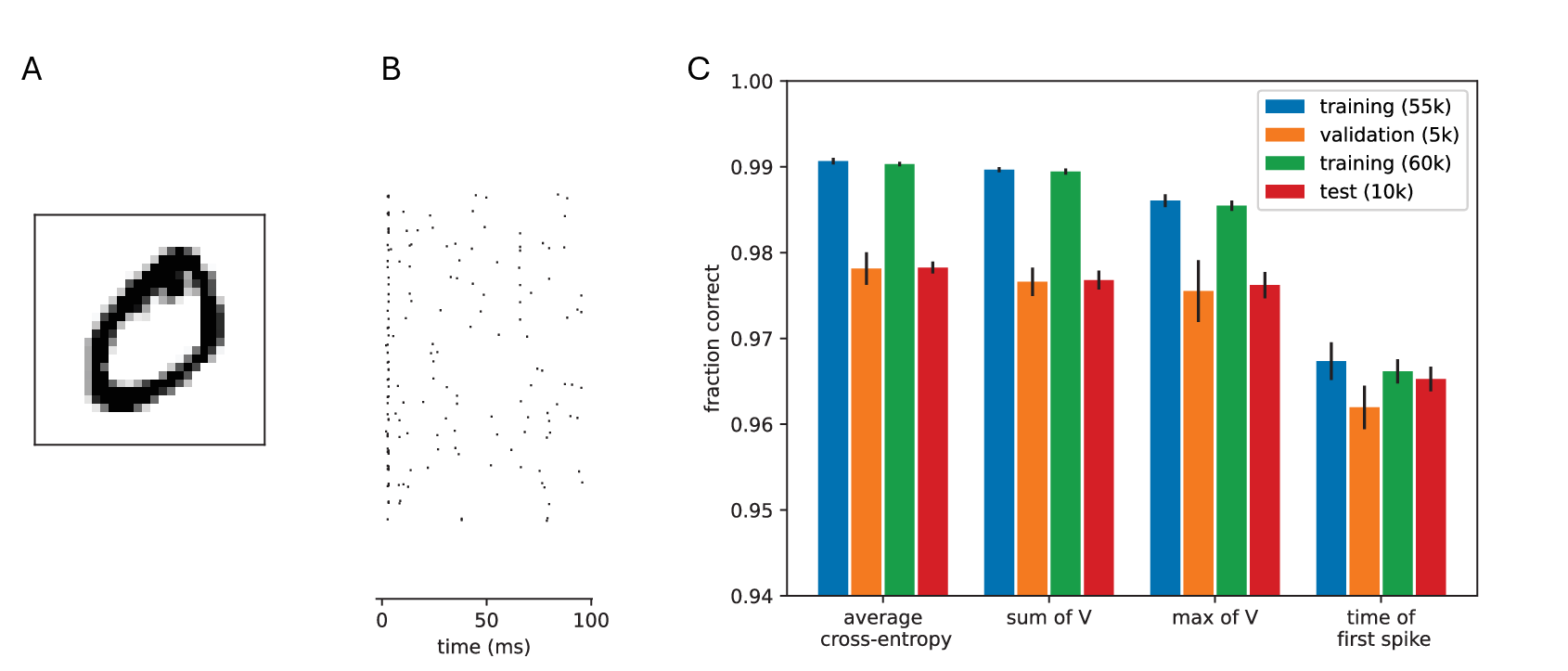}
   \caption{Overview of the classification performance of a feedforward Eventprop network on the latency encoded MNIST dataset. {\bf A)} Example $28 \times 28$ digit from the MNIST dataset. {\bf B)} The image from A encoded in the spike latencies $t_i$ of neurons $i= 0,\ldots,783$, using the formula $t_i= \frac{255-x_i}{255} (T_{\text{trial}} - 4$ ms$)+2$ ms, where $x_i$ is the grey level of the $i$th pixel in a row-wise translation through the MNIST image. {\bf C)} Classification accuracy with different loss functions. Bars are the mean of $10$ independent runs and errorbars the standard deviation. All training was for $50$ epochs. There was no regularisation in the hidden layer except for when ${\cal L}_{\text{time}}$ was used. In this case $k_{\text{reg}}= 10^{-8}$ and $\nu_{\text{hidden}}= 4$. }
    \label{fig:MNIST_overview}
\end{figure}

\clearpage

\begin{landscape}

\begin{figure}
    \noindent $\!\!\!\!\!$
    \includegraphics[width=23cm]{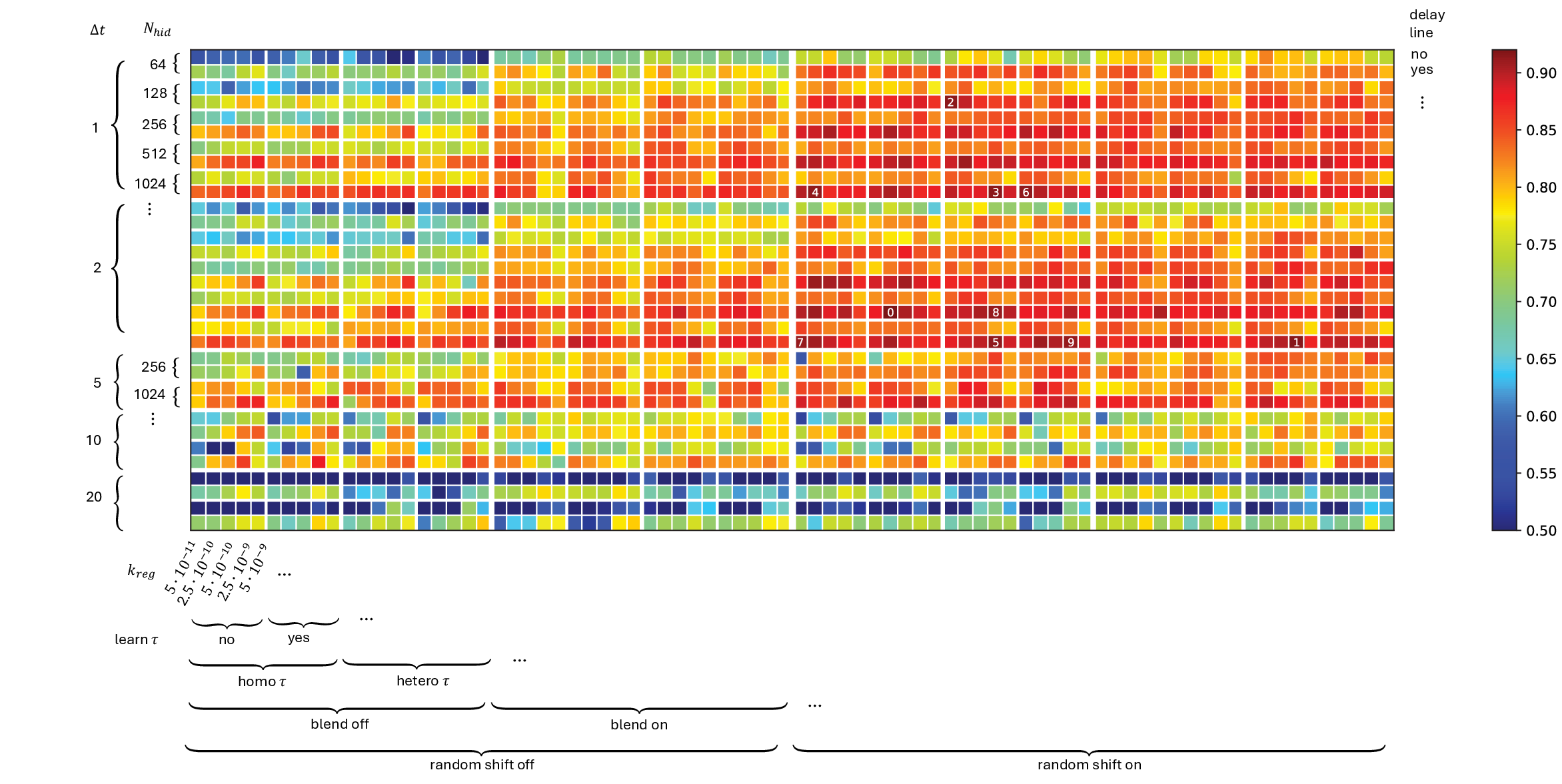}
    \caption{Overview of the final validation accuracy during 10-fold leave-one-speaker-out cross-validation on the SHD dataset. The colours code the observed validation accuracy at the epoch of best achieved training error, averaged across all 10 folds and across 2 independent runs with different seeds. The positions of the coloured cells indicate the parameter settings as indicated on the axes. The numbers 0 to 9 in white indicate the 10 best-performing parameter combinations according to this measurement. The colour range is from $0.5$ to the maximal observed values to highlight differences between the well-performing networks.\label{fig:SHD_final_valid_map}}
\end{figure}

\begin{figure}
    \noindent $\!\!\!\!\!$
    \includegraphics[width=23cm]{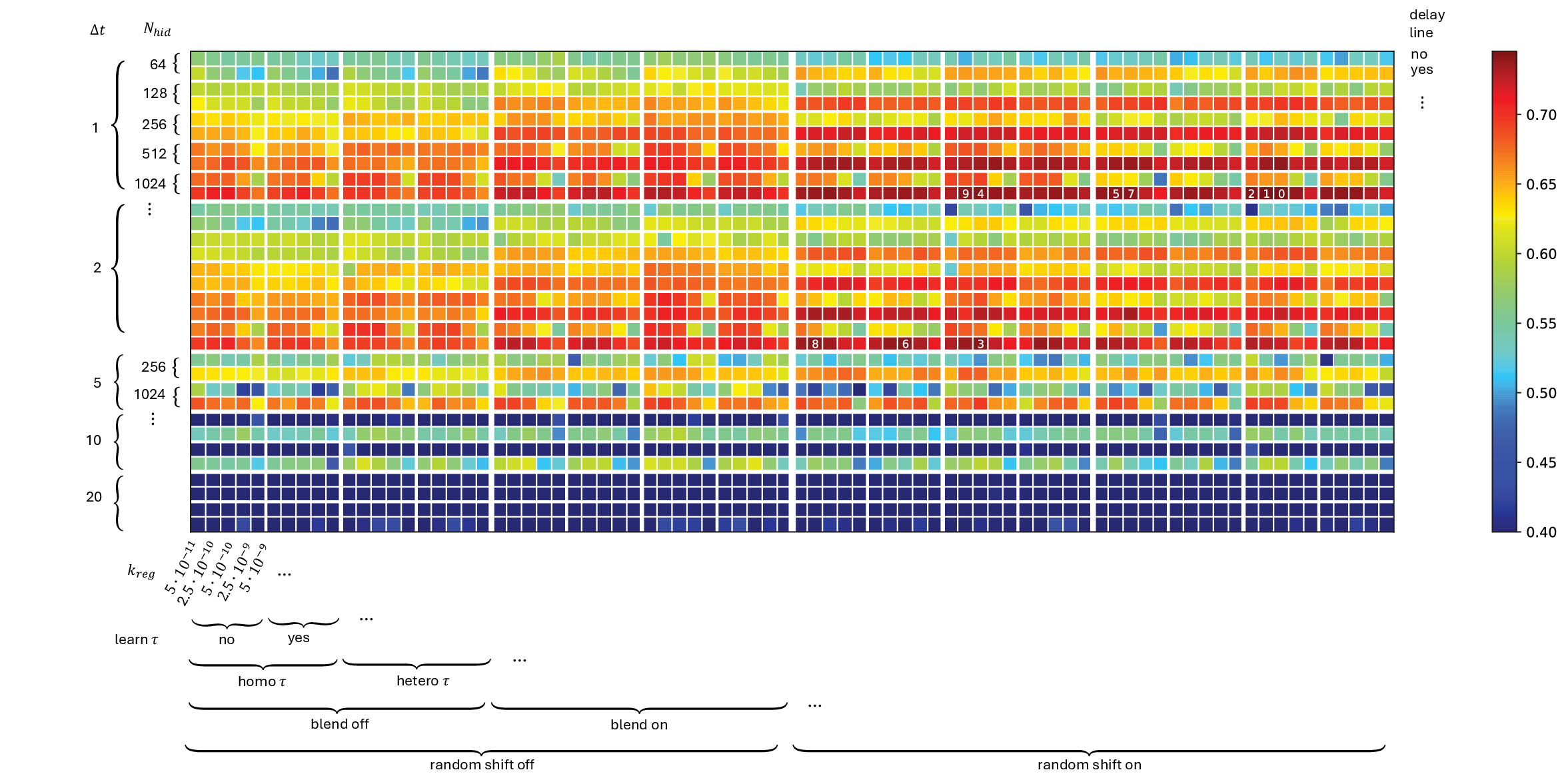}
    \caption{Overview of the best validation accuracy observed on the SSC validation set. The colours code the validation accuracy at the epoch where it is maximal, averaged across 2 independent runs with different seeds. The positions of the coloured cells indicate the parameter settings and the numbers 0 to 9 in white the best networks, as in figure \ref{fig:SHD_final_valid_map}. The colour range is from $0.4$ to the maximal observed values.\label{fig:SSC_best_valid_map}}
\end{figure}

\end{landscape}

\end{document}